\documentclass[10pt,a4paper]{article}

\usepackage{xcolor}
\usepackage[utf8]{inputenc}
\usepackage{amsmath,amssymb,amsthm}
\usepackage{framed}
\usepackage{a4wide}
\usepackage{hyperref}
\usepackage{booktabs}
\usepackage{array}
\usepackage{tabularx}
\usepackage{tikz}
\usepackage{natbib}

\usepackage{authblk}

\definecolor{shadecolor}{gray}{0.95}
\definecolor{flowblue}{RGB}{59,105,160}
\definecolor{flowgreen}{RGB}{72,135,104}
\definecolor{floworange}{RGB}{190,118,45}
\definecolor{flowgray}{RGB}{246,247,249}
\newcolumntype{Y}{>{\raggedright\arraybackslash}X}
\usetikzlibrary{arrows.meta,positioning,fit}

\newtheorem{theorem_inner}{Theorem}[section]

\newtheorem{proposition_inner}[theorem_inner]{Proposition}

\begin{document}

\title{Alignment and Safety of Diffusion Models via Reinforcement Learning and Reward Modeling: A Survey}

\author[1]{Preeti Lamba}
\author[1]{Kiran Ravish}
\author[1]{Ankita Kushwaha}
\author[1]{Pawan Kumar} 
\affil[1]{International Institute of Information Technology, Hyderabad, India}

\maketitle

\begin{abstract}
Diffusion models have become a central paradigm for image and multimodal generation, yet their deployment raises persistent questions about alignment, safety, preference satisfaction, and robustness to misuse. This survey reviews recent progress on aligning text-to-image diffusion models through reinforcement learning, reward modeling, preference optimization, and safety-specific fine-tuning. We organize the literature along five axes: the source of feedback, the form of the reward or preference signal, the optimization mechanism, the treatment of distribution shift and reward overoptimization, and the extent to which safety is addressed as an explicit constraint rather than a generic preference. The review covers reinforcement learning from human feedback, KL-regularized policy optimization, direct preference optimization, binary utility optimization, differentiable reward fine-tuning, surrogate reward learning, region-aware fine-tuning, and safety-oriented DPO variants. To make the survey accessible, we include tutorial explanations of diffusion sampling, reward modeling, and preference optimization, and briefly connect image diffusion alignment to emerging text and masked language diffusion models. We also compare representative methods in terms of feedback requirements, computational cost, scalability, susceptibility to reward hacking, and suitability for safety-critical deployment. Finally, we synthesize the literature into a set of open challenges: multi-objective alignment, feedback-efficient preference learning, adversarially robust safety alignment, continual alignment under changing norms, and interpretable reward modeling. The goal of this survey is to provide a coherent technical map of the emerging area of diffusion model alignment and to identify the methodological gaps that must be addressed before aligned generative models can be reliably deployed.
\end{abstract}


\tableofcontents

\section{Introduction}

Generative diffusion models have achieved remarkable success in producing high-quality images from text prompts, demonstrating state-of-the-art results in text-to-image synthesis \cite{ho2020ddpm,dhariwal2021diffusion,nichol2021improved,rombach2022high,nichol2022glide,saharia2022imagen}. Models such as DALL-E, Imagen, and Stable Diffusion represent a paradigm shift in how AI can create content, with applications ranging from art generation to data augmentation. Despite their power, these models are typically trained on broad internet data via maximum likelihood objectives, which do not necessarily align model outputs with what humans consider desirable or safe \cite{deng2024prdp,prabhudesai2023alignprop}. As a result, unaligned diffusion models may produce content that is mismatched with user intent (e.g., failing to follow certain prompt details \cite{lee2023aligning}), offensive or unsafe (due to learned biases or harmful data in training sets \cite{yang2024dense}), or otherwise not in line with human aesthetic preferences \cite{xu2023imagereward}.

Alignment of generative models refers to steering model behavior toward human values, preferences, and safety constraints. In natural language processing, significant progress has been made by techniques like reinforcement learning from human feedback (RLHF) \cite{ouyang2022training}. InstructGPT-style systems demonstrated that supervised instruction tuning followed by human-preference reward modeling and policy optimization can improve helpfulness and safety behavior \cite{christiano2017deep}. These successes have motivated analogous efforts to align image generation models. However, aligning diffusion models poses unique challenges: the output is high-dimensional (images), feedback can be expensive (requiring human comparisons of images), and direct optimization is complicated by the iterative denoising process of diffusion \cite{yang2024dense}.

This survey focuses on \textbf{reinforcement learning and reward modeling approaches to alignment in diffusion models}. In this setting, \emph{reward modeling} denotes the construction of a proxy objective that reflects human preferences, prompt fidelity, aesthetic quality, or safety constraints, while \emph{reinforcement learning} treats the diffusion sampling process as a policy whose parameters can be updated using scalar feedback. Recent work spans a broad methodological spectrum:

\begin{itemize}
    \item Using human feedback to train reward models for images (e.g., aesthetic or relevance scores) \cite{xu2023imagereward} and then applying RL or related optimization to fine-tune the diffusion model to increase those rewards \cite{fan2023dpok}.
    \item \textit{Reinforcement learning from human feedback (RLHF)} for images: collecting human preference data (e.g., which of two images better matches a prompt) and using policy gradient methods (or variants) to adjust the model \cite{lee2023aligning}.
    \item \textit{Direct Preference Optimization (DPO)} methods: optimizing the model directly on preference data without an explicit reward model by turning preference comparisons into a loss \cite{rafailov2023dpo}.
    \item \textit{Differentiable reward fine-tuning}: techniques that backpropagate reward gradients through the diffusion sampling process, avoiding high-variance RL estimators \cite{clark2024draft}.
    \item Addressing safety-specific alignment: methods focusing on reducing harmful content while preserving quality \cite{xing2025focusfix}, for example by optimizing against a ``safety reward'' or filtering mechanism.
\end{itemize}

Despite these efforts, many open problems remain. Early studies showed that naive fine-tuning on human feedback can lead to \emph{alignment--fidelity trade-offs} \cite{lee2023aligning}: improving prompt alignment may degrade image fidelity or diversity. Over-optimizing against a learned reward model can cause \emph{reward hacking}, where the generative model finds loopholes in the reward that do not correspond to genuinely better images \cite{yang2024dense,zhang2024rewardoveropt}. Ensuring \textbf{safety} (avoiding nudity, violence, or bias) is particularly challenging; it often requires additional precautions since simply instructing the model via text may not suffice. Feedback data can also be expensive to gather and may not cover the full space of prompts and user intents, raising questions of generalization and scalability \cite{xu2023imagereward,uehara2024feedback}.

This review has three objectives. First, it provides a structured taxonomy of alignment methods for diffusion models, emphasizing approaches based on reinforcement learning, reward models, preference data, and differentiable rewards. Second, it compares these methods according to their optimization objectives, feedback assumptions, sample efficiency, stability, and safety implications. Third, it identifies open problems that remain insufficiently addressed by the current literature, including reward overoptimization, multi-objective trade-offs, adversarial prompt robustness, continual preference drift, and interpretability of reward models.

The remainder of the paper is organized as follows. Section~\ref{sec:background} reviews diffusion models, language-diffusion variants, and the alignment problem. Section~\ref{sec:literature} surveys major families of alignment methods, including RLHF-style approaches, reward modeling, direct preference optimization, differentiable reward fine-tuning, and safety-specific methods. Section~\ref{sec:related} analyzes methodological relationships among these works and highlights how the field has evolved from reward-model-based RL toward more direct, efficient, and safety-aware objectives. Section~\ref{sec:research} synthesizes the survey into a set of open research challenges. Section~\ref{sec:conclusion} concludes with a discussion of the main lessons and unresolved issues.

\section{\label{sec:background}Background: Diffusion Models and Alignment Problem}

\subsection{Diffusion Models for Generative AI}
Diffusion models are a class of generative models that learn to synthesize data by reversing a noising process. Originally developed as Denoising Diffusion Probabilistic Models (DDPMs) \cite{ho2020ddpm}, they have become a dominant approach for image generation. In a typical text-to-image diffusion model (e.g., Stable Diffusion), the model starts from random noise and iteratively denoises it to produce an image, guided by a text prompt. Formally, a diffusion model defines a forward process that gradually adds noise to an image $x_0$ over $T$ steps, producing latent $x_T \sim \mathcal{N}(0,I)$, and a learned reverse model $p_\theta(x_{t-1}\mid x_t, c)$ that predicts the previous step given the current noisy image $x_t$ and condition $c$ (such as text) \cite{ramesh2022hierarchical}. After training on a large image-text dataset to approximate the true distribution of images conditioned on text, the model can generate high-fidelity images for new prompts.

For a new reader, the key idea is simple: training deliberately corrupts real data, and generation learns to undo that corruption. In the forward process, a clean sample $x_0$ is gradually transformed into noise through known transitions $q(x_t\mid x_{t-1})$. In the reverse process, a neural network learns to predict either the previous sample $x_{t-1}$, the original clean sample $x_0$, or the noise that was added at step $t$. Once trained, sampling starts from noise $x_T$ and repeatedly applies the learned denoiser until a clean sample is obtained. Conditioning information, such as a text prompt $c$, is injected into the denoiser so that the reverse process produces images compatible with the prompt.

\begin{figure}[t]
\centering
\begin{tikzpicture}[
    font=\small,
    box/.style={draw=flowblue, rounded corners=2pt, thick, fill=flowgray, align=center, minimum width=2.3cm, minimum height=0.9cm},
    model/.style={draw=flowgreen, rounded corners=2pt, thick, fill=flowgray, align=center, minimum width=3.0cm, minimum height=0.9cm},
    reward/.style={draw=floworange, rounded corners=2pt, thick, fill=flowgray, align=center, minimum width=2.6cm, minimum height=0.85cm},
    arrow/.style={-Latex, thick}
]
\node[box] (x0) {clean data\\$x_0$};
\node[box, right=0.9cm of x0] (xt) {partly noised\\$x_t$};
\node[box, right=0.9cm of xt] (xT) {near noise\\$x_T$};
\draw[arrow] (x0) -- node[above=13pt]{add noise} (xt);
\draw[arrow] (xt) -- node[above=13pt]{add noise} (xT);

\node[model, below=1.0cm of xt] (denoiser) {learned denoiser\\$\epsilon_\theta(x_t,t,c)$ or $p_\theta(x_{t-1}\mid x_t,c)$};
\node[reward, below=1.0cm of denoiser] (align) {alignment signal\\reward, preference, or safety score};

\draw[arrow] (xT.south) |- node[pos=0.25,right]{reverse sampling} (denoiser.east);
\draw[arrow] (denoiser.west) -| node[pos=0.25,below=4pt]{denoise} (x0.south);
\draw[arrow] (align) -- node[right]{fine-tuning signal} (denoiser);
\end{tikzpicture}
\caption{A diffusion model is trained by adding noise to data and learning a reverse denoising process. Alignment methods add an extra signal, such as a reward model or preference loss, that changes the denoiser so that generated samples better match human intent or safety constraints.}
\label{fig:diffusion-tutorial}
\end{figure}

The common notation is as follows. The variable $x_0$ denotes clean data, $x_t$ denotes a partially corrupted sample at time $t$, $x_T$ is nearly pure noise, and $c$ is optional conditioning information. The model parameters are denoted by $\theta$. In image diffusion, $x_t$ may be pixels or latent variables; latent diffusion models such as Stable Diffusion perform denoising in a compressed latent space rather than directly in pixel space, which reduces computational cost while preserving high visual quality \cite{rombach2022high}.

The success of diffusion models in image generation is evident in their photorealism and diversity of outputs \cite{ho2020ddpm,dhariwal2021diffusion,song2021scorebasedgenerativemodelingstochastic,ramesh2022dalle2,rombach2022high}. However, because they are trained on broad data with a simple objective (usually a noise prediction loss related to maximizing image likelihood), these models often reflect undesired aspects of their training corpus:
\begin{itemize}
    \item \textbf{Prompt Misalignment}: The model sometimes fails to accurately fulfill all aspects of the user's prompt (e.g., getting object counts or colors wrong) \cite{lee2023aligning}. This occurs because the model optimizes for likelihood, not necessarily semantic faithfulness to the prompt (especially for rare or complex prompt conditions).
    \item \textbf{Undesired Content}: They may output explicit or violent imagery if prompted, or even without explicit prompting, due to learning from such content in training data. Biases present in the training data (e.g., stereotypes) can also manifest in outputs.
    \item \textbf{Quality Issues}: The model may generate low-quality, blurry, or artifact-ridden images under some conditions, particularly if the user's notion of quality (sharpness, aesthetics) is not explicitly part of the training objective \cite{xu2023imagereward}.
\end{itemize}

Addressing these issues is crucial for practical deployment. Users typically want models that not only produce realistic images, but also follow instructions accurately and refrain from disallowed content. This is where the notion of alignment comes into play.

\subsection{Text and Masked Diffusion Models}
Although this survey focuses mainly on text-to-image diffusion alignment, diffusion modeling is also an active direction for language generation. The central difficulty is that text is discrete: a token is a category, not a continuous pixel value. Early approaches handled this by diffusing continuous word embeddings, as in Diffusion-LM, which iteratively denoises Gaussian vectors into word vectors and supports gradient-based control for text generation \cite{li2022diffusionlm}. DiffuSeq applied diffusion to sequence-to-sequence generation and showed that diffusion models can be competitive on conditional text generation tasks while producing diverse outputs \cite{gong2023diffuseq}.

A second line of work uses discrete or masked corruption processes. D3PMs introduced structured discrete denoising transitions, including absorbing or masking states, and connected diffusion models to autoregressive and mask-based generation \cite{austin2021structured}. SEDD extended score-matching ideas to discrete data through score entropy and reported strong language-modeling performance relative to prior diffusion language models \cite{lou2024sedd}. Masked Diffusion Language Models (MDLMs) showed that a simple masked-discrete objective, implemented as a mixture of masked language modeling losses, can substantially narrow the gap between diffusion and autoregressive language modeling \cite{sahoo2024mdlm}. At larger scale, LLaDA trains a Transformer-based masked diffusion model from scratch: a forward process masks tokens, and the reverse process predicts masked tokens in parallel while simulating generation from fully masked text to complete text \cite{nie2025llada}. This differs from standard autoregressive LLMs, which generate strictly left-to-right.

Preference optimization for masked language diffusion follows the same high-level goal as image alignment: increase the probability of preferred responses while constraining drift from the base model. The technical challenge is different because a masked diffusion model often uses an evidence lower bound (ELBO) or a stochastic estimate of sequence likelihood rather than an exact left-to-right log probability. LLaDA 1.5 introduced Variance-Reduced Preference Optimization (VRPO), which analyzes the variance of ELBO-based likelihood estimates and uses Monte Carlo budget allocation and antithetic sampling to reduce gradient variance during preference optimization \cite{zhu2025llada15}. This illustrates a broader point: preference optimization transfers across modalities, but each generative family requires estimators that respect its own likelihood or sampling structure.

\subsection{Alignment and Safety in Generative Models}
We define \textbf{alignment} as the process of modifying a model such that its outputs and behavior adhere to specified human preferences or values. In the context of diffusion models, this could mean:
\begin{itemize}
    \item Ensuring the generated image closely matches the user's intent (as described by the text prompt). 
    \item Optimizing images for human-preferred qualities like aesthetic appeal \cite{xu2023imagereward} or realism.
    \item Avoiding outputs that are offensive, unsafe, or violate policy (e.g., hate symbols, gore, NSFW content).
\end{itemize}
\textbf{Safety} is a subset of alignment focusing on preventing harmful outcomes. A safely aligned diffusion model should refuse or alter prompts aiming to produce disallowed content, and generally reduce the presence of bias or toxicity in its output.

The alignment problem can be viewed as introducing \emph{additional objectives} on top of the original generative modeling goal. The base model learns $p_\theta(x\mid c)$ for images $x$ given prompt $c$. The alignment objective is to adjust $\theta$ so that it also maximizes a \textit{reward function} $R(x,c)$ that captures alignment goals. For example, $R$ may be high if the image is photorealistic and matches the prompt, or low if the image is unsafe. In language models, $R$ is often given by a learned preference model or by human judgments.

However, directly incorporating these objectives into diffusion models is non-trivial. If $R(x,c)$ is not differentiable or given implicitly (e.g., via human comparisons), standard training has to be adapted. Moreover, the diffusion model's output is generated through many intermediate steps $x_T \to \dots \to x_0$, which complicates credit assignment: did a misalignment issue arise from an early step or a later step?

\paragraph{Reinforcement Learning Formulation:} We can conceptualize the diffusion sampling process as a Markov Decision Process (MDP) where:
\begin{itemize}
    \item The \textit{state} can be the current diffusion step $x_t$.
    \item The \textit{action} is the choice of denoising step to $x_{t-1}$.
    \item The \textit{policy} $\pi_\theta$ corresponds to the diffusion model's conditional distribution at each step.
    \item A \textit{reward} $R$ is obtained at the end of the diffusion (only at $t=0$, when a complete image is generated) based on how aligned the result is.
\end{itemize}
This MDP is unusual because the policy is already a strong pretrained diffusion model and the reward is sparse, appearing only at the end of sampling. Standard RL algorithms could, in principle, fine-tune $\pi_\theta$ to maximize expected reward $\mathbb{E}_{x\sim \pi_\theta}[R(x,c)]$ for prompts $c$. Several works have studied policy-gradient-style updates for diffusion models in this setting \cite{fan2023dpok}; PRDP \cite{deng2024prdp} does not rely on policy gradients, but it analyzes related stability issues.

Another perspective is \textbf{reward-weighted regression} or \textbf{preference optimization}. Given a dataset of outputs with preference labels (e.g., $A$ is better than $B$ for a given prompt), one can update the model to increase the likelihood of preferred outputs over non-preferred ones \cite{xu2023imagereward}. This can be done without explicit RL credit assignment by deriving a loss function from preferences, as in DPO for language \cite{rafailov2023dpo} and its diffusion-model adaptations.

\paragraph{Reward Modeling:} Often, the true objective of interest (human satisfaction) is hard to measure automatically. Reward modeling addresses this by training a \textit{reward model} $r_\phi(x,c)$ to approximate human judgments. For images, this may involve collecting a dataset of human comparisons (which of two images is better for the prompt) and fitting $r_\phi$ so that $r_\phi(x^+,c) > r_\phi(x^-,c)$ for preferred image $x^+$ over $x^-$. Once $r_\phi$ is learned, it can provide a dense, differentiable reward signal for any generated image. This model can then guide diffusion-model alignment, either through RL (treating $r_\phi$ as the reward function in the RL objective) or direct gradients when possible \cite{clark2024draft}.

A tutorial view of RLHF is shown in Figure~\ref{fig:rlhf-tutorial}. The base model first generates candidate outputs. Human annotators, or a trusted evaluator, compare candidates and indicate which one better satisfies the prompt and policy. These comparisons train a reward model. The generator is then updated to increase reward while staying close to the original model, usually through a KL penalty or another proximity constraint. Direct preference methods such as DPO remove the explicit reward-model stage and train directly from preferred/dispreferred pairs, but they still rely on the same core information: which output is better under human or policy preferences.

\begin{figure}[t]
\centering
\begin{tikzpicture}[
    font=\small,
    stage/.style={draw=flowblue, rounded corners=2pt, thick, fill=flowgray, align=center, minimum width=2.55cm, minimum height=0.9cm},
    opt/.style={draw=flowgreen, rounded corners=2pt, thick, fill=flowgray, align=center, minimum width=2.8cm, minimum height=0.9cm},
    arrow/.style={-Latex, thick}
]
\node[stage] (gen) {generate\\candidates};
\node[stage, right=0.75cm of gen] (pref) {collect\\preferences};
\node[stage, right=0.75cm of pref] (rm) {train reward\\model $r_\phi$};
\node[opt, right=0.75cm of rm] (opt) {optimize model\\with KL control};
\draw[arrow] (gen) -- (pref);
\draw[arrow] (pref) -- (rm);
\draw[arrow] (rm) -- (opt);
\draw[arrow] (opt.north) to[out=100,in=80,looseness=0.65] node[above]{new model generates better candidates} (gen.north);
\node[draw=floworange, rounded corners=2pt, thick, fill=flowgray, align=center, below=1.35cm of pref, minimum width=3.4cm] (dpo) {DPO-style methods:\\optimize from pairs directly};
\draw[arrow] (pref) -- (dpo);
\draw[arrow] (dpo.east) -| (opt.south);
\end{tikzpicture}
\caption{RLHF-style alignment trains a reward model from preferences and then optimizes the generator against that reward while limiting drift. DPO-style methods use the same pairwise preference information but fold the reward-model step into a direct training objective.}
\label{fig:rlhf-tutorial}
\end{figure}

For example, \textit{ImageReward} is a reward model learned from human ratings of images given prompts \cite{xu2023imagereward}. Such a model can score any candidate image. Using $r_\phi$ as a proxy, one can fine-tune the generator to increase $r_\phi$ scores of its outputs \cite{lee2023aligning}. However, caution is needed: $r_\phi$ is an imperfect proxy and optimizing too hard against it can cause the generator to exploit quirks of $r_\phi$ (reward hacking). This phenomenon, also called \textit{reward overoptimization}, is analogous to an agent exploiting a mis-specified reward function in RL, often at the expense of the intended outcome \cite{yang2024dense,zhang2024rewardoveropt}.

In summary, aligning diffusion models involves:
\begin{itemize}
    \item Defining or learning a reward that captures alignment goals (human preference or safety).
    \item Fine-tuning or guiding the diffusion model to optimize this reward, while avoiding degeneration (maintaining diversity, preventing reward hacking, etc.).
    \item Doing so efficiently (with as little human data as possible) and robustly (generalizing to new prompts, not just those seen during feedback collection).
\end{itemize}
In the next section, we review the literature that has tackled these tasks, categorizing approaches by how they use reward models and reinforcement learning techniques.

\section{\label{sec:literature}Literature Review of Alignment Methods for Diffusion Models}
A number of key approaches to align text-to-image diffusion models with human preferences have been proposed in recent years. We structure this review along methodological lines:
\begin{enumerate}
    \item \textbf{Reinforcement Learning with Human Feedback (RLHF) for Diffusion Models}: Methods that explicitly apply RL algorithms (e.g., policy gradient, PPO) using a reward model or human feedback signal.
    \item \textbf{Direct Preference Optimization (DPO) and Variants}: Methods that bypass training a reward model by directly using preference data (human choices) in the loss function, often yielding a simpler fine-tuning procedure.
    \item \textbf{Differentiable Reward Fine-Tuning}: Techniques that assume a differentiable reward function (or make one via surrogate models) and directly backpropagate into the diffusion model, avoiding RL's high variance.
    \item \textbf{Advanced Strategies to Mitigate Issues}: Hybrid or improved methods addressing specific challenges such as reward overoptimization, sample inefficiency, and safety constraints, including reward difference modeling \cite{deng2024prdp,zhang2024aligningfewstepdiffusionmodels} and region-specific tuning \cite{xing2025focus}.
    \item \textbf{Feedback Efficiency and AI Feedback}: Methods that reduce reliance on human feedback by being more sample-efficient or by using AI-generated feedback \cite{lee2024rlaifvsrlhfscaling} (RLAIF). In diffusion-model alignment, such feedback may come from vision-language models rather than LLMs alone \cite{bordes2024introductionvisionlanguagemodeling}, although the quality and bias of the feedback model remain central concerns.
\end{enumerate}

Table~\ref{tab:methods} at the end of this section provides a high-level comparison of representative methods across these categories. We now discuss each category with key examples from the literature.

\subsection{RLHF for Diffusion: Policy Gradient Approaches}
The direct application of reinforcement learning to diffusion model alignment typically involves the following pipeline: (1) gather human feedback on model outputs, such as rankings by quality or prompt alignment; (2) train a reward model on this data to predict a scalar reward; and (3) fine-tune the diffusion model with an RL algorithm to maximize this reward model's output. This mirrors the approach used in NLP for aligning LMs \cite{ouyang2022training}.

\textbf{Aligning Text-to-Image Models using Human Feedback} \cite{lee2023aligning} was one of the first such attempts. In this work, the authors collected human feedback for a variety of prompts, focusing on image-text alignment issues (such as counting objects correctly). They trained a reward model $r_\phi$ on these human labels. Then, instead of a full RL algorithm, they used a simpler ``reward-weighted likelihood'' fine-tuning: essentially re-weighting the diffusion model's training loss by $\exp(r_\phi)$. This can be seen as a form of RL where the gradient approximates an advantage-weighted update. Their results showed better prompt fidelity (e.g., correctly coloring objects as instructed) after fine-tuning \cite{lee2023aligning}. They also analyzed trade-offs, noting that pushing alignment too far can reduce image diversity or quality. This work laid groundwork on design choices (how to incorporate the reward, how to balance it with the original loss, etc.) and showed that even a relatively straightforward RLHF method can significantly improve alignment for specific attributes.

\textbf{DPOK (Fan et al., NeurIPS 2023)} \cite{fan2023dpok} took a more rigorous RL approach. The authors define an online RL fine-tuning procedure for diffusion models with a reward model. DPOK uses \emph{policy gradient with KL-regularization}. In practice, this is analogous to the PPO algorithm used in language model RLHF, where the diffusion model (policy) is updated to increase reward but is penalized for deviating too much from the original model (to maintain image quality and diversity). The ``K'' in DPOK indicates the inclusion of a KL term. They also provide analysis on the role of this KL regularization in stability \cite{fan2023dpok}. Experiments showed DPOK outperformed supervised fine-tuning baselines in both alignment (e.g., better prompt compliance) and quality metrics \cite{fan2023dpok}. DPOK can be seen as an evolution of the approach by Lee et al. \cite{lee2023aligning}, introducing a more formal RL objective and constraints to mitigate unwanted drift. It still relies on a pretrained reward model to guide the optimization.

A challenge in these RLHF approaches is the \textbf{sparse reward}: the model gets a single score after an image is fully generated. This is akin to a bandit setting with no intermediate feedback. As diffusion models typically sample many steps, RL gradients such as REINFORCE can have high variance. One strategy is to use \emph{rejection sampling or reranking}: generate several images per prompt and select the one with the highest reward \cite{lee2023aligning}. This improves outputs without changing the model, but it increases inference cost and does not permanently update the generator. RL fine-tuning addresses this by updating the model itself to more often produce high-reward images on the first try.

Several works have reported that naive RLHF, such as using REINFORCE directly on diffusion outputs, can be unstable or require many samples \cite{gupta2025simpleeffectivereinforcementlearning}. To tackle this, alternative RL formulations have been explored:
\textbf{Feedback-Efficient Online RL (Uehara et al., ICML 2024)}: Instead of offline feedback and a static reward model, Uehara et al. consider an online setting where the model queries an oracle for the reward of generated samples and updates continuously \cite{uehara2024feedback}. They frame it as a bandit problem on the space of generated data, with the goal of finding high-reward regions of the image distribution efficiently \cite{uehara2024feedback}. They propose an RL algorithm that biases exploration toward feasible samples, since many random images may be nonsensical or out-of-distribution \cite{uehara2024feedback}. By focusing on the manifold of realistic images, the method improves the sample efficiency of finding better samples. The theoretical contribution is a regret bound showing that the algorithm converges to high-reward outputs with fewer queries \cite{uehara2024feedback}. This is important for scalability because it suggests ways to fine-tune models with substantially less human evaluation. While their examples include images (aesthetics) and molecules, the method is a general RL approach. One limitation is that it assumes an oracle can provide a reward for any sample on demand; for molecules this may be a property calculator, whereas for images it may require a human-in-the-loop process that is difficult to scale.

\textbf{Handling Reward Hacking (Zhang et al., ICML 2024)}: Reward overoptimization occurs when the diffusion model exploits weaknesses in the reward model rather than improving the intended output quality. Zhang et al. analyze this phenomenon in diffusion models and identify two biases \cite{zhang2024rewardoveropt}: (1) a temporal inductive bias, because diffusion models operate sequentially and earlier denoising steps establish structure that later steps refine; and (2) a primacy bias in reward-model neurons, where saturated critic neurons become easier to exploit while dormant neurons can act as regularizers \cite{zhang2024rewardoveropt}. Based on these observations, they proposed \textbf{TDPO-R (Temporal Diffusion Policy Optimization with neuron Reset)} to modify the RL algorithm \cite{zhang2024rewardoveropt}. TDPO introduces a temporally aware objective and includes a mechanism to reset or penalize over-active critic neurons during training \cite{zhang2024rewardoveropt}. This changes the gradient seen by the diffusion model and discourages it from exploiting narrow reward-model features. Empirically, TDPO-R achieved better human-rated outputs for the same reward gain, indicating reduced reward hacking relative to baseline RL methods \cite{zhang2024rewardoveropt}.

More recently, several works have adapted \textbf{Group Relative Policy Optimization (GRPO)} to image and video generation. GRPO is attractive because it avoids training a separate value critic: for a prompt $c$, the model samples a group of $K$ outputs, scores them with a reward model, and converts their within-group relative rewards into advantages. In diffusion, each output corresponds to a full denoising trajectory, so the policy update can be viewed as increasing the probability of high-reward denoising trajectories while constraining the model to remain close to the pretrained generator. Figure~\ref{fig:grpo-tutorial} summarizes this loop.

\begin{figure}[t]
\centering
\begin{tikzpicture}[
    font=\small,
    block/.style={draw=flowblue, rounded corners=2pt, thick, fill=flowgray, align=center, minimum width=2.05cm, minimum height=0.85cm},
    rewardblock/.style={draw=floworange, rounded corners=2pt, thick, fill=flowgray, align=center, minimum width=2.25cm, minimum height=0.85cm},
    updateblock/.style={draw=flowgreen, rounded corners=2pt, thick, fill=flowgray, align=center, minimum width=2.3cm, minimum height=0.85cm},
    arrow/.style={-Latex, thick}
]
\node[block] (prompt) {prompt\\$c$};
\node[block, right=0.45cm of prompt, yshift=0.82cm] (traj1) {trajectory 1\\$x_T\!\to\!x_0^{(1)}$};
\node[block, right=0.45cm of prompt] (traj2) {trajectory 2\\$x_T\!\to\!x_0^{(2)}$};
\node[block, right=0.45cm of prompt, yshift=-0.82cm] (trajk) {trajectory $K$\\$x_T\!\to\!x_0^{(K)}$};
\node[rewardblock, right=0.55cm of traj2] (score) {reward model\\$r_1,\ldots,r_K$};
\node[updateblock, right=0.5cm of score] (adv) {group-relative\\advantages $A_i$};
\node[updateblock, right=0.5cm of adv] (upd) {policy update\\KL controlled};

\draw[arrow] (prompt.east) -- (traj1.west);
\draw[arrow] (prompt.east) -- (traj2.west);
\draw[arrow] (prompt.east) -- (trajk.west);
\draw[arrow] (traj1.east) -- (score.west);
\draw[arrow] (traj2.east) -- (score.west);
\draw[arrow] (trajk.east) -- (score.west);
\draw[arrow] (score) -- (adv);
\draw[arrow] (adv) -- (upd);
\draw[arrow] (upd.north) to[out=120,in=60,looseness=0.75] node[above]{next group} (prompt.north);
\end{tikzpicture}
\caption{GRPO-style alignment samples a group of denoising trajectories for the same prompt, scores the final outputs, normalizes rewards within the group to obtain relative advantages, and updates the generator with a KL or reference-model constraint.}
\label{fig:grpo-tutorial}
\end{figure}

\textbf{DanceGRPO (Xue et al., 2025)} \cite{xue2025dancegrpo} presents a unified GRPO framework for visual generation across diffusion models and rectified flows, covering text-to-image, text-to-video, and image-to-video tasks. It evaluates on models including Stable Diffusion, HunyuanVideo, FLUX, and SkyReel-I2V, using reward models for image/video aesthetics, text-image alignment, video motion quality, and binary feedback \cite{xue2025dancegrpo}. The main contribution is showing that GRPO can provide stable online RL alignment at a scale where earlier RL methods such as DDPO and DPOK can become unstable.

\textbf{Flow-GRPO (Liu et al., 2025)} \cite{liu2025flowgrpo} focuses on flow-matching models, where deterministic ODE sampling makes exploration less direct than in stochastic diffusion sampling. It introduces an ODE-to-SDE conversion so that RL can explore stochastic trajectories while preserving the original marginal distributions, and a denoising reduction strategy to lower training cost without changing the number of inference steps \cite{liu2025flowgrpo}. This makes GRPO-style online RL applicable to modern flow-based text-to-image generators, with reported gains on compositional generation, visual text rendering, and human preference alignment.

Subsequent GRPO variants refine efficiency and credit assignment. \textbf{BranchGRPO} restructures rollouts into a branching tree so that common prefixes are shared, sparse terminal rewards are converted into denser depth-wise advantages, and low-value or redundant paths can be pruned \cite{li2025branchgrpo}. \textbf{AEGPO} uses attention-entropy signals to allocate rollout budgets across prompts and to focus exploration on denoising timesteps where attention is most dispersed \cite{li2026aegpo}. Together, these works show a trend from simply applying policy gradients to diffusion trajectories toward designing rollout, exploration, and advantage estimators that respect the structure of image generation.

In summary, RLHF applied to diffusion models has proven effective but comes with challenges of stability and potential misalignment due to the proxy reward. Methods like DPOK, TDPO-R, DanceGRPO, and Flow-GRPO add regularization, group-relative baselines, or diffusion/flow-specific inductive biases to keep online RL stable. RLHF methods typically require a reward model or evaluator trained on human feedback, discussed next.

\subsection{Reward Modeling and Preference Data for Diffusion}
Before discussing direct preference optimization methods, it is useful to review efforts to create \textbf{reward models and datasets} for image generation alignment.

\begin{itemize}

    \item \textbf{ImageReward (Xu et al., NeurIPS 2023)}: This work introduced a large-scale dataset of human comparisons for text-to-image outputs and trained a reward model on it \cite{xu2023imagereward}. The authors collected 137k comparisons of images generated by various models given the same prompt, labeled by human raters for preference \cite{xu2023imagereward}. The resulting \textit{ImageReward model} predicts a score given an image and prompt. It was shown to correlate better with human preferences than prior automatic metrics \cite{heusel2017gans,li2022blip,radford2021learning,schuhmann2022laion}. ImageReward enabled researchers to have a ready-to-use $r_\phi(x,c)$ for experiments, which likely accelerated RLHF research for diffusion models. Xu et al. in \cite{xu2023imagereward} also proposed a simple fine-tuning approach called \textit{ReFL (Reward Feedback Learning)} to optimize a diffusion model against the ImageReward score \cite{xu2023imagereward}. This method, similar in spirit to RLHF, demonstrated that tuning to maximize the learned reward improved human-rated quality of outputs in their tests.
    
    \item \textbf{Better Aligning T2I with Human Preference (Wu et al., ICCV 2023) \cite{wu2023humanpreferencescorev2}}: This paper took a somewhat different angle, focusing on improving the \emph{evaluation} of alignment and constructing a new benchmark called \textit{HPSv2 (Human Preference Score v2) \cite{wu2023humanpreferencescorev2}}. Wu et al. collected a dataset of user preferences (798,090 human preference choices on 433,760 pairs of images) and trained an evaluator model. They fine-tuned diffusion models (particularly Stable Diffusion \cite{rombach2022high}) and observed that certain training schemes improved the alignment score without heavy RL. While they did not introduce a fundamentally new RL method, their contribution in data and metrics (like proposing \textit{PickScore}, an automatic metric learned from the Pick-a-Pic dataset) provided tools for later works to measure success. This addresses the need for reliable evaluation: since running a human study for each new method is costly, having a stand-in metric like PickScore or ImageReward to at least approximate human judgments is useful.
    \item \textbf{Pick-a-Pic (Kirstain et al., NeurIPS 2023)}: Although primarily a dataset paper, Pick-a-Pic released an open dataset of user preference judgments for images \cite{kirstain2023pickapic}. It contains tens of thousands of prompt-image pairs with binary feedback (which image is preferred). Such datasets are key for both training reward models and directly training diffusion models on preferences.
\end{itemize}

These efforts underline that \textbf{reward modeling is a critical component} of alignment. A good reward model enables alignment algorithms to progress without constant human intervention. However, training a robust reward model is itself challenging: it must generalize to outputs from a model after it is fine-tuned (distribution shift) and not encode unintended biases. Some research has looked into more general reward-model representations, for example by using CLIP-based image features.

In the language domain, \textbf{Direct Preference Optimization (DPO)} \cite{rafailov2023dpo} avoids training an explicit reward model and directly fine-tunes a policy so that $\pi_\theta(\text{output}^+)$ is higher than $\pi_\theta(\text{output}^-)$ for preferred versus dispreferred outputs. This yields a simple cross-entropy-style loss on binary comparisons, avoiding explicit reinforcement learning. The key insight is to treat the base model as a prior and derive a Bayes-optimal update from pairwise feedback, resulting in a loss $\mathcal{L} \propto -\log \sigma(\beta (f_\theta(x^+) - f_\theta(x^-)))$, where $f_\theta = \log \pi_\theta$ is the log-probability up to a constant and $\beta$ is a temperature hyperparameter \cite{rafailov2023dpo}.

For diffusion models, applying DPO is more complicated than for language because we need a notion of $\pi_\theta(x|c)$ for images. Diffusion models define a distribution over the image (through the iterative process). In principle, one could treat the final sample $x_0$ probability under the model as $\pi_\theta(x_0|c)$, but calculating or differentiating that is intractable due to the latent variable of the diffusion chain. However, some works have adapted preference optimization ideas:

\textbf{Diffusion-DPO (Miao et al., 2023)} \cite{wallace2023diffusionmodelalignmentusing}: Building upon Direct Preference Optimization (DPO), traditionally used in language models, the authors adapt this approach for diffusion models by incorporating a differentiable objective derived from the evidence lower bound. Utilizing the Pick-a-Pic dataset, which contains over 851,000 pairwise human preferences, they fine-tuned the Stable Diffusion XL (SDXL)-1.0 model. The resulting model demonstrated significant improvements in visual appeal and prompt alignment compared to both the base SDXL-1.0 and its larger variant with an additional refinement model. Additionally, the study explores a variant of Diffusion-DPO that employs AI-generated feedback, achieving performance levels comparable to those trained with human preferences, suggesting a scalable approach to diffusion model alignment.

The reported results in Diffusion-KTO \cite{li2024diffusionkto} (below) suggest that a straightforward adaptation of DPO to diffusion is possible but might not be optimal for heavily distilled models.

\textbf{D3PO (Yang et al., CVPR 2024)} \cite{yang2024d3po}: ``Direct Preference for Denoising Diffusion Policy Optimization'' explicitly tackled the challenge of preference-based fine-tuning without a reward model. The authors identify memory issues that arise when preferences are incorporated across multi-step generation. D3PO introduces a multi-step objective that decomposes the preference signal across diffusion timesteps \cite{yang2024d3po}. It treats each diffusion step as part of an expanded MDP, so that outcome-level preferences can be integrated into timestep-level updates while reducing memory requirements. D3PO fine-tunes a model using only human comparisons, achieving effects similar to RLHF while avoiding reward-model training. A notable advantage of not using a reward model is avoiding reward-model mis-specification. However, one trade-off is that preference data must be available for direct fine-tuning, whereas a reward model can generalize to new prompts or be reused across multiple fine-tuning rounds. The results of D3PO were competitive, particularly in reducing memory requirements relative to naive RL \cite{yang2024d3po}.

\textbf{Aligning Diffusion with Human Utility (Li et al., NeurIPS 2024)} \cite{li2024diffusionkto}: The Diffusion-KTO approach can be seen as another form of preference optimization. Instead of pairwise comparisons, it assumes binary per-image feedback, such as user likes or dislikes. It then formulates the fine-tuning objective as maximizing expected utility, which under binary feedback means increasing the probability of liked images. This approach does not need a reward model or pairwise comparisons; it uses a large volume of weak feedback that can be obtained from implicit user behavior or ratings \cite{li2024diffusionkto}. The authors showed that using such signals, their method outperformed both supervised fine-tuning and an implementation of Diffusion-DPO \cite{li2024diffusionkto}. This implies that readily available data, such as user ratings or curated good-versus-bad image sets, can be useful without curated preference comparisons. Diffusion-KTO optimizes $\theta$ to maximize $P_\theta(\text{``user likes image''}\mid c)$ across the training distribution of prompts and user feedback. It uses a loss resembling logistic regression on binary feedback, treating the diffusion model as producing a score for each image.

\textbf{Personalized preference optimization} extends these ideas beyond a single population-level reward. \textbf{Personalized Preference Fine-tuning of Diffusion Models (PPD)} \cite{dang2025personalizedpreference} learns user preference embeddings from a small number of pairwise examples using a vision-language model, injects those embeddings into the diffusion model through cross-attention, and fine-tunes with a DPO objective conditioned on the user embedding. This lets one model represent multiple users' preferences and generalize to unseen users with only a few examples \cite{dang2025personalizedpreference}. For personalized editing, \textbf{Collaborative DPO (C-DPO)} represents users as nodes in a preference graph, learns user embeddings with a lightweight graph neural network, and optimizes edits with a DPO-style objective that balances individual alignment with neighborhood coherence among users with similar visual tastes \cite{dunlop2025personalizedediting}. These works are important because they clarify that ``human preference'' is not always a single scalar target: two users can reasonably prefer different color palettes, styles, or editing outcomes for the same prompt.

In summary, DPO and related preference-based methods simplify the pipeline by removing the intermediate reward model. They often produce more stable training because cross-entropy losses are well behaved, and they are computationally cheaper because they do not require iterative RL rollouts. However, they typically require feedback in a particular format, such as direct rankings of outputs from the current model distribution. If the model changes substantially during training, a fixed comparison dataset may become less relevant; this is also a concern in RLHF if the reward model is not updated. Some approaches mitigate this issue by iteratively collecting preferences, although image-domain preference collection still often requires human input.

Direct preference methods also underscore an important concept: \textit{the diffusion model itself can serve as its own reward model to some extent}. For example, DPO in language pointed out the connection between model logits and reward. In diffusion, some works, such as \cite{zhu2025dspo}, draw connections between maximizing reward and adjusting the model's score function directly. This leads naturally into the next category.

\subsection{Differentiable Reward Fine-Tuning (Backpropagation-Based)}
One way to avoid the complexity of RL is if the reward function is differentiable with respect to the image, and ideally, if we can differentiate through the generation process. In other words, instead of treating the diffusion model as a black-box policy to be nudged via RL, treat the whole sampling as a computation graph and do gradient ascent on reward.

This direction was advanced by two contemporaneous works:

\begin{itemize}
    \item \textbf{DRaFT: Direct Reward Fine-Tuning (Clark et al., ICLR 2024)} \cite{clark2024draft}: The authors showed that one can backpropagate through the entire diffusion sampling process, from noise $x_T$ to image $x_0$, to obtain the gradient of a reward $R(x_0)$ with respect to model parameters. The diffusion process consists of $T$ stochastic sampling steps, but the method uses differentiable sampling and reparameterization ideas to propagate reward gradients through the denoising trajectory \cite{Kingma_2019}. DRaFT also introduced two approximations: truncating backpropagation to the last $K$ steps (DRaFT-K), and a lower-variance estimator for the single-step case (DRaFT-LV). These approximations keep memory and computation manageable. Experiments on Stable Diffusion fine-tuning for rewards such as CLIP score and aesthetic score showed that DRaFT achieved higher reward faster than RL-based methods. The work also connects several algorithms under a shared view, showing how RL, classifier guidance, and other techniques relate mathematically.
    
    \item \textbf{AlignProp (Prabhudesai et al., 2023)} \cite{prabhudesai2023alignprop}: Similarly, AlignProp proposed end-to-end backpropagation of the reward gradient through the diffusion model. The focus in AlignProp was engineering: storing all gradients is memory-heavy, so it fine-tunes only small \emph{adapter modules} (low-rank adapter layers inserted in the model) and uses gradient checkpointing to drop intermediate results and recompute them as needed. This makes the gradient computation fit in memory. AlignProp was tested on multiple reward objectives, including semantic alignment, aesthetics, and controllability such as object count. Because it can optimize any differentiable reward, it can combine rewards by summing them in the loss. The results showed that differentiable rewards can substantially improve sample efficiency. Qualitatively, if the reward is to contain exactly two objects, AlignProp can adjust the model to reliably output images with exactly two instances of the subject, a fine-grained control that is difficult for naive diffusion models.
\end{itemize}

The core idea in both is treating the diffusion model's sampling as differentiable. In practice, we have to deal with the stochastic sampling (which is why gradient checkpointing or reparameterization is needed). These methods essentially do gradient descent on $-\mathbb{E}_{x_0 \sim p_\theta}[R(x_0)]$ with respect to $\theta,$ by swapping expectation and gradient under assumptions or approximations.

One limitation is that the reward must be differentiable or approximated by a differentiable surrogate. A human preference is not directly differentiable, but a neural network proxy such as ImageReward or CLIP score is. Thus, these methods often still rely on a learned reward model. The difference is that they avoid the RL step and directly update the generator according to the reward model's gradient.

Directly optimizing a reward model can also be risky when the reward model is imperfect. This is the reward hacking scenario: AlignProp \cite{prabhudesai2023alignprop} and DRaFT \cite{clark2024draft} can overfit to the reward model if run too long. Regularization, such as a KL penalty that keeps the aligned model close to the original model, or early stopping is used to mitigate this risk \cite{wallace2023diffusionmodelalignmentusing,liu2024alignmentdiffusionmodelsfundamentals}. In RL terms, this corresponds to limiting movement in policy space.

Interestingly, \textbf{Reward Surrogate Optimization} has been explored to allow differentiable training for \emph{non-differentiable rewards}. A prime example:
\textbf{LaSRO (Jia et al., CVPR 2025)} \cite{jia2025lasro}, or Latent Space Reward Optimization, targets \emph{two-step diffusion models}, including highly distilled models that generate in one or two steps. Traditional RL can struggle in this setting because the process is highly non-Markovian over a very small number of large denoising steps \cite{smith2024diffusiondpo}. LaSRO learns a differentiable \textit{surrogate reward} in the latent space of the diffusion model \cite{lee2024diffusionkto}. For example, if the true reward is a user preference between images A and B, LaSRO trains a surrogate reward function $\tilde{R}(z)$, where $z$ is the latent representation of an image, so that $\tilde{R}(z)$ approximates the user preference. It leverages a pretrained latent diffusion model so that $z$ captures image content \cite{zhang2024surrogate}. Because the surrogate is differentiable with respect to $z$, and $z$ is part of the generation process, the method can backpropagate through $\tilde{R}$ and the generator. This converts a potentially non-differentiable reward into a learned differentiable reward in a lower-dimensional space. With this approach, LaSRO fine-tuned a two-step diffusion model and outperformed RL baselines, including implementations of DDPO and Diffusion-DPO, while improving stability \cite{jia2025lasro}.
 
Differentiable fine-tuning methods have clear advantages in speed and often in final performance because they use gradient information more directly. Their main disadvantage is reliance on reward-model quality and differentiability. In cases where the reward is a true black box, such as direct human yes/no feedback without a learned proxy, RL remains a fallback. In many practical settings, however, a neural reward model can be fit to human feedback and then used by differentiable methods.

\subsection{Safety-Specific and Region-Specific Alignment Techniques}
While most of the above works optimize for generic ``preferences'' or ``aesthetics,'' some focus specifically on \textbf{safety alignment}. Safety can be framed as just another reward to optimize (e.g., a reward model that detects unsafe content and gives low score), but it has some unique aspects:
\begin{itemize}
    \item There are hard constraints: certain content must never appear, rather than merely being lower preference.
    \item Mitigations should ideally not drastically alter images otherwise considered good.
    \item The space of ``unsafe prompts'' or cases can be different from normal usage distribution.
\end{itemize}

An example in this category is \textbf{AlignGuard (2024)} \cite{liu2024safetydposcalablesafetyalignment}. It applies preference-style training to safety alignment by collecting pairwise data in which one image is safer than another and directly optimizing this preference signal. Using the CoProV2 dataset, the method trains specialized ``safety experts'' in the form of Low-Rank Adaptation (LoRA) matrices. These experts guide the model's generation process away from specific harmful concepts. AlignGuard adopts a DPO-style loss on safety-labeled pairs.

A novel approach to minimizing negative side effects is \textbf{Focus-N-Fix (Xing et al., CVPR 2025)} \cite{xing2025focus}. Instead of applying a global fine-tune for safety or quality, Focus-N-Fix identifies \textit{problematic regions} in generated images, such as regions that cause low reward or safety violations, and fine-tunes the model to change only those regions. The idea follows from the observation that fine-tuning a model on one aspect, such as reducing violent content, can inadvertently affect prompt fidelity or style. Focus-N-Fix is a region-aware algorithm that uses an attention mask during fine-tuning to restrict updates to portions of the image identified as problematic. For example, if an output image contains an over-sexualized region detected by an NSFW classifier, Focus-N-Fix adjusts how the model generates that region without altering the rest of the composition. It does so by augmenting the diffusion process with region information, guiding the model to treat flagged pixels differently. Experiments showed improvements in localized metrics, such as reducing unsafe content while preserving background and other details. This approach is promising for safety because it targets specific failure regions rather than altering the model's entire distribution. Across many such cases, the model learns to avoid problematic patterns with little or no degradation in other quality metrics.

Another safety-relevant direction uses \textbf{self-critique models or AI feedback}. Some RLAIF-style works train reward models not from human data but from another AI system, such as a safety classifier, an LLM, or a vision-language model that provides feedback on image content. This can provide large-scale feedback without direct human annotation, but it also risks inheriting the feedback model's biases and errors.

\subsection{Summary of Methods and Comparative View}
Having reviewed categories of methods, we now summarize and compare key dimensions in Table~\ref{tab:methods}. The dimensions chosen are:
\begin{itemize}
    \item \textbf{Method and Reference}: identifying the approach.
    \item \textbf{Alignment Objective}: what type of reward or preference it optimizes (e.g., human preference via reward model, aesthetic score, binary feedback, etc.).
    \item \textbf{Technique}: RL (and which algorithm) vs. direct optimization vs. other.
    \item \textbf{Feedback Modality}: human comparisons, human ratings, AI-generated rewards, etc.
    \item \textbf{Efficiency/Scalability}: notes on whether the method is sample-efficient or requires large feedback datasets, and how it scales to high-resolution or many prompts.
    \item \textbf{Safety Considerations}: whether the method explicitly addresses safety and how (e.g., through a safety-specific reward or by design).
\end{itemize}

\begin{table}
\tiny
\centering
\caption{Comparison of Alignment Methods for Diffusion Models}
\label{tab:methods}
\begin{tabularx}{\textwidth}{>{\raggedright\arraybackslash}p{1.8cm} Y Y Y Y Y}
\toprule
\textbf{Method (Year)} & \textbf{Alignment Objective} & \textbf{Technique} & \textbf{Feedback Type} & \textbf{Efficiency/Scale} & \textbf{Safety Features} \\
\midrule
Lee et al. 2023 \cite{lee2023aligning} & Prompt-image alignment (accuracy on attributes) & Reward-weighted likelihood (implicit RL) & Human-labeled comparisons (custom dataset) & Medium (requires new human data, moderate size) & No explicit safety, general alignment focus \\
DPOK (2023) \cite{fan2023dpok} & Human preference reward (image-text match, quality) & Online RL (policy grad + KL) & Human feedback via reward model & Low (needs many samples, but KL improves stability) & Not specifically, aims not to degrade any aspect \\
ImageReward + ReFL (2023) \cite{xu2023imagereward} & Aesthetic and similarity preferences (general-purpose) & Reward model + direct fine-tune (some RL-like steps) & Large human comp dataset (137k comps) & High (pre-trained reward, one can reuse it; fine-tune efficient) & Indirect (if safety included in pref, otherwise no) \\
AlignProp (2023) \cite{prabhudesai2023alignprop} & Differentiable rewards (CLIP, aesthetic, etc.) & Backprop through diffusion (low-rank adapters) & Any differentiable feedback (often learned models) & High (very fast convergence, needs reward model) & Can include safety classifier gradient as reward \\
DRaFT (2024) \cite{clark2024draft} & Differentiable rewards (often learned) & Backprop (full or truncated) & Same as above & High (less memory with truncation, fast) & If a safety reward model used, yes; method itself agnostic \\
PRDP (2024) \cite{deng2024prdp} & Learned reward model (black-box) & Reward Difference Regression + RL (proximal updates) & Human pref via reward model & High (stable on 100K+ prompts, scalable to large data) & Possibly addressed indirectly by stability (no specific safety metric) \\
TDPO-R (2024) \cite{zhang2024rewardoveropt} & Human reward model & Policy gradient (PPO-like) with temporal & Human reward model & Medium (focused on avoiding pitfalls, not on reducing data) & Yes, mitigates reward hacking which includes extreme outputs \\
Feedback-Efficient RL (2024) \cite{uehara2024feedback} & Task-specific ground-truth reward (e.g., aesthetic or molecule property) & Online RL exploration (theory-backed) & Oracle feedback (human or simulator) & High (fewer queries for high reward) & Could be used for safety by treating safe output as reward \\
DanceGRPO (2025) \cite{xue2025dancegrpo} & Image/video preference rewards & GRPO over denoising trajectories & Reward models or binary feedback & High (group-relative baseline improves online RL stability) & Safety possible if reward includes safety \\
Flow-GRPO (2025) \cite{liu2025flowgrpo} & T2I compositionality, OCR, and preference rewards & GRPO for flow matching via ODE-to-SDE conversion & Reward models & High (denoising reduction lowers training cost) & Not explicit, but compatible with safety rewards \\
D3PO (2024) \cite{yang2024d3po} & Human preferences (pairwise) & Direct preference optimization (multi-step) & Human comparisons (no reward model) & Medium (needs preferences for current model outputs, alleviates memory issues of naive approach) & Not specifically addressed \\
Dense Reward View (2024) \cite{yang2024dense} & Human preferences & DPO-style with temporal discount & Human comparisons (like DPO, but with time-aware loss) & Medium (similar to DPO, slightly more overhead) & Not specifically, but helps general alignment \\
Diffusion-KTO (2024) \cite{li2024kto} & Binary per-image feedback (likes) & Utility maximization (logistic objective) & Implicit human feedback (e.g., from user behavior) & High (uses abundant weak feedback, no custom labeling needed) & Can integrate safety by treating only safe outputs as liked \\
PPD (2025) \cite{dang2025personalizedpreference} & User-specific preferences & Personalized DPO with user embeddings & Few-shot pairwise preferences & High (one model can represent multiple users) & Can personalize safety/style preferences if data supports it \\
C-DPO (2025) \cite{dunlop2025personalizedediting} & Personalized image-editing preferences & Collaborative DPO with graph user embeddings & User preferences plus neighbor signals & Medium (uses preference graph structure) & Not safety-specific \\
DSPO (2025) \cite{zhu2025dspo} & Human preferences (or style, etc.) & Direct Score Preference Optimization & Preference pairs & High (optimizes the diffusion score model directly) & Not explicitly, but applicable if preferences involve safety \\
LaSRO (2025) \cite{jia2025lasro} & Arbitrary rewards (including non-diff) for 2-step model & Learned latent reward (differentiable) + gradient & Human or other (converted to latent surrogate) & Medium-High (solves previously unsolvable 2-step case, requires surrogate training) & Focuses on preserving quality in ultra-fast models (no direct safety metric) \\
BranchGRPO (2025) \cite{li2025branchgrpo} & Image/video preference rewards & Tree-structured GRPO with branching and pruning & Reward models & High (shares prefixes and improves credit assignment) & Not explicit, but compatible with safety rewards \\
AEGPO (2026) \cite{li2026aegpo} & T2I preference rewards & Entropy-guided GRPO rollout allocation & Reward models & High (prioritizes informative prompts and timesteps) & Not explicit, but compatible with safety rewards \\
Focus-N-Fix (2025) \cite{xing2025focusfix} & Localized quality (safety, artifacts) & Region-based fine-tuning (mask problematic) & Human or automated region annotations (e.g. detector highlights) & High (does not need huge data, each finetune targeted, can generalize) & Yes, explicitly targets safety issues (over-sexualization, violence) region-wise \\
AlignGuard (2024) \cite{liu2024safetydposcalablesafetyalignment} & Safety compliance preferences & DPO-style training on safety-labeled pairs & Human or curated safety comparisons & Medium (similar to DPO) & Yes, entire objective is safety alignment \\
\bottomrule
\end{tabularx}
\end{table}

From the above comparison, a few trends can be observed:

There is a clear trade-off between methods requiring extensive human data (e.g., RLHF with large custom datasets) and those leveraging more readily available or simulated signals (binary feedback, AI rewards). Approaches like Diffusion-KTO \cite{li2024diffusionkto} and PRDP \cite{wang2024prdp} strive to be scalable to web-scale feedback or prompt sets.

Early RLHF methods established effectiveness but newer methods like AlignProp and DRaFT tend to dominate in terms of efficiency, often achieving equal or better alignment in less time \cite{clark2024draft,prabhudesai2023alignprop}.

Ensuring safety often requires special handling; a generic ``improve preference'' method might not inherently prevent all bad outputs unless the reward model was trained to include that in its scoring. Thus, approaches like Focus-N-Fix or AlignGuard emerged to improve safety without sacrificing alignment on other axes \cite{xing2025focus,liu2024safetydposcalablesafetyalignment}.

Many methods are complementary. For instance, ImageReward can be used as the reward model in AlignProp or DRaFT, while PRDP's stable objective can be followed by Focus-N-Fix to address safety-specific details. The field is converging toward hybrid designs: differentiable methods for speed, RL-style constraints to avoid overoptimization, and region- or timestep-aware mechanisms for specific failure modes.

Having reviewed the literature, the next section highlights connections between these works and how they build on each other.

\section{\label{sec:related}Inter-Paper Relationships and Methodological Progress}
The rapid development of alignment techniques for diffusion models has been highly interrelated. We now outline some key relationships and evolutionary trends.

\subsection{From RLHF to Direct Optimization}
The progression from \cite{lee2023aligning} and \cite{fan2023dpok} (RL with human feedback) to \cite{prabhudesai2023alignprop} and \cite{clark2024draft} (direct backpropagation) illustrates a shift in strategy. Early successes in RLHF confirmed that human feedback can significantly improve outputs \cite{lee2023aligning}, but also revealed issues such as instability and high-variance gradients. AlignProp and DRaFT respond to these issues by computing exact or approximate reward gradients and by leveraging reward models, such as ImageReward \cite{xu2023imagereward}, more efficiently than rollout-based RL.

Once a reward model such as ImageReward \cite{xu2023imagereward} became available, the bottleneck shifted to using it effectively. AlignProp reports that vanilla RL (REINFORCE) is sample-inefficient and slow, motivating reward backpropagation \cite{prabhudesai2023alignprop}. DRaFT similarly shows significant speedups over an RL baseline for the same reward \cite{clark2024draft}. Thus, one line of progress is to obtain a strong reward model through preference data, then replace rollout-based RL with gradient-based fine-tuning for efficiency.

\subsection{Incorporating Inductive Biases of Diffusion}
Several papers build on the fact that diffusion models are not generic policies: they have a multi-step denoising structure and a specific forward/reverse process. For example, the \textbf{Dense Reward View (Yang et al.)} \cite{yang2024dense} highlights the temporal aspect. Yang et al. take inspiration from DPO but argue that a diffusion model's multi-step nature violates the i.i.d. assumption behind the original DPO derivation \cite{yang2024dense}. By adding temporal discount factors to DPO's loss, the method prioritizes early denoising steps, consistent with the view that early steps encode coarse composition and later steps add detail \cite{yang2024dense}. This was partially motivated by Zhang et al.'s observation that ignoring temporal bias can cause reward overoptimization \cite{zhang2024rewardoveropt}.

TDPO (Temporal DPO) explicitly uses temporal inductive bias to modify training \cite{yang2024dense}, while related work on primacy bias introduces critic-neuron resetting to reduce reward-model exploitation \cite{zhang2024rewardoveropt}. These developments show how ideas from language-model RLHF, such as avoiding reward hacking, are adapted to image generation with diffusion-specific temporal structure.

Focus-N-Fix's region-based idea is another inductive bias: images have spatial structure, and misalignment may be spatially localized \cite{xing2025focus}. Instead of treating the image as only an output vector, it uses computer vision tools to identify problematic regions and adjust only those regions. This is reminiscent of classic computer vision approaches that segment or localize problematic content; Focus-N-Fix brings that localization into the model training loop. For localization, current foundation models such as SAM \cite{kirillov2023segment} can be useful.

\subsection{Shared Use of Data and Models}
Many of these works rely on common datasets or models, showing how one paper's contribution enables others:

The \textbf{Human Preference Dataset (HPS)} \cite{wu2023humanpreferencescorev2} and \textbf{Pick-a-Pic} \cite{kirstain2023pickapic} appear in multiple works. PRDP used Human Preference Dataset v2 and Pick-a-Pic in large-scale training \cite{wang2024prdp}. This illustrates how shared preference datasets enable evaluation and algorithm development at scale.

\textbf{ImageReward model} is cited widely as a baseline or component. AlignProp's demo comparisons use HPS v2 \cite{wu2023humanpreferencescorev2} and ImageReward \cite{xu2023imagereward} as benchmarks. Diffusion-KTO \cite{li2024diffusionkto} uses ImageReward and PickScore as automatic evaluation metrics to claim better performance. This indicates a convergence on certain standard reward models (ImageReward, PickScore) for evaluation, similar to how language models often use HELM \cite{liao2025helmhumanpreferredexplorationlanguage} or other curated metrics to compare alignment.

Because of these shared references, the community can make incremental comparisons: if PRDP shows that RL can work at $100K$ scale and yields certain output quality \cite{wang2024prdp}, then a method like DRaFT can attempt the same scale and evaluate improvement. DRaFT's results report improvements over RL-based approaches on a variety of rewards \cite{clark2024draft}, using methods such as DPOK as comparisons.

\subsection{Complementary Nature of Approaches}
While at first glance some methods compete (e.g., RL vs direct optimization), they can be combined.
\begin{itemize}
    \item \textbf{Two-Stage Training}: A direct optimization method can quickly reach a useful alignment regime, after which RL can incorporate non-differentiable feedback or fresh human judgments on the updated model's outputs.

    \item \textbf{Regularization Mix}: KL penalties from RLHF can be included inside gradient-based methods, for example as a term in AlignProp's loss that keeps the aligned model close to the original model. Conversely, RL methods such as DPOK already include KL regularization as a core design. Across method families, there is convergence around constraining distribution shift to prevent quality collapse.

    \item \textbf{Safety + Preference}: Focus-N-Fix can be applied after other alignment processes. For example, a model fine-tuned with AlignProp for aesthetic quality and prompt alignment may still contain localized unsafe patterns. Running Focus-N-Fix with a safety reward can address those regions locally. Because Focus-N-Fix reports safety improvements with little or no degradation in other aspects \cite{xing2025focus}, it can serve as a safety-specific stage layered on top of a generally aligned model.
\end{itemize}

\subsection{Incremental Improvements}

Some works explicitly build atop previous ones.

PRDP (CVPR 2024) \cite{deng2024prdp} followed DPOK (NeurIPS 2023) \cite{fan2023dpok} and addressed scalability issues by changing the objective to reward difference prediction \cite{wang2024prdp}. In this sense, DPOK established RLHF-style fine-tuning for diffusion models, while PRDP developed a more scalable version of the same broad direction.

Diffusion-KTO (NeurIPS 2024) \cite{li2024diffusionkto} introduced a complementary direction based on binary feedback. This broadens the range of usable supervision by showing that weaker feedback signals can be competitive with pairwise comparisons in some settings.

\subsection{Competitive Benchmarks}

As these methods matured, they increasingly compared against one another:

\begin{itemize}
    \item AlignProp and DRaFT compare direct reward backpropagation against RL-style baselines, while DSPO studies preference optimization directly at the level of diffusion score functions \cite{prabhudesai2023alignprop,clark2024draft,zhu2025dspo}.
    \item Diffusion-KTO compares against Diffusion-DPO and reports improvements in human judgment and ImageReward scores \cite{li2024diffusionkto}.
    \item PRDP compares against established RL-based methods in small-scale settings and shows that PRDP scales more reliably to larger settings \cite{wang2024prdp}.
    \item Many papers use baseline models such as Stable Diffusion v1.4 or v1.5 and evaluate on common prompt sets, which makes incremental improvements more measurable.
\end{itemize}

In conclusion, the papers form a web of progress: initial attempts adapted the known RLHF idea to images (with some success but limitations), then others improved the efficiency (AlignProp, DRaFT), stability (PRDP, TDPO), and specificity (Focus-N-Fix, AlignGuard) of those methods. Data and evaluation tools produced by one were reused by others. The community is moving towards more sample-efficient and safe alignment techniques, with overlapping ideas reinforcing the validity of certain approaches (for example, multiple groups found direct gradient methods effective, which validates that direction).

The literature review and analysis inform the future work: despite all the progress, open problems remain. The next section proposes five such problems and research directions to address them.

\section{\label{sec:research}Open Problems and Research Directions}
Drawing from the survey above, we identify five open challenges for alignment and safety in diffusion models. Each challenge is described in terms of its significance, key technical obstacles, and relation to the reviewed literature.

\textit{How can diffusion models be aligned simultaneously to multiple human preferences, such as semantic accuracy, aesthetic quality, and safety, without sacrificing one objective for another?} Most existing works optimize a single reward at a time or sum different aspects into one score, which can lead to trade-offs where improving one aspect, such as safety, may degrade another, such as prompt fidelity \cite{clark2024draft,prabhudesai2023alignprop}. This motivates methods for \textbf{multi-objective alignment} that ensure a model satisfies several criteria at once.

\paragraph{Key Challenges:} 

\begin{itemize}
    \item \textbf{Reward Balancing}: Different objectives (say, matching prompt vs. avoiding offensive content) have rewards on different scales and frequencies. A naive weighted sum of rewards might still allow the model to game one at expense of others (e.g., achieving high aesthetic score by introducing minor prompt errors).
    
    \item \textbf{Catastrophic Forgetting}: Fine-tuning on one objective and then on another can make the model forget the first. If we do sequential alignment (first aesthetic, then safety), the first goal might regress. Joint training is needed, which is tricky if objectives conflict.
    
    \item \textbf{Evaluation}: Success must be measured on multiple axes, either through a combined metric or by visualizing a Pareto frontier where improving one objective may hurt another.
    
    \item \textbf{Human Preference Trade-offs}: Ultimately, human users might accept slight losses in one area for gains in another (e.g., a tiny drop in image sharpness is fine if safety is ensured). Capturing these trade-offs either in a reward model or in the training procedure is hard.
\end{itemize}

\paragraph{Related Work:} 
Several works examine trade-offs directly. Lee et al. \cite{lee2023aligning} analyze alignment--fidelity trade-offs and show that careful hyperparameter choices are needed. Focus-N-Fix \cite{xing2025focusfix} addresses related concerns by localizing changes to avoid global degradation. AlignProp \cite{prabhudesai2023alignprop} and DRaFT \cite{clark2024draft} show that differentiable rewards can be combined by addition, but this assumes that a linear combination is adequate.

In RL, scalarization and Pareto-optimal policy search provide foundations for multi-objective optimization, but they have not yet been fully adapted to diffusion-model alignment. Conditional training is another possible direction: a model could be trained with a control variable that adjusts the trade-off between safety strictness and image variety, analogous to how classifier-free guidance adjusts fidelity and prompt adherence.

\textit{Can the amount of human feedback required to align a diffusion model be reduced through active learning or synthetic feedback generation?} The cost of obtaining human preference data is one of the largest bottlenecks for image-domain RLHF. Current methods often assume a static dataset of comparisons or ratings, although only a fraction may be maximally informative for improving the model. This motivates systems that query humans selectively and use \textbf{AI feedback} to filter or predict less critical cases.

\paragraph{Key Challenges:}
\begin{itemize}
    \item \textbf{Uncertainty Quantification}: The model needs to estimate when human preferences are uncertain. Diffusion models do not naturally provide uncertainty over preference; reward-model uncertainty, such as preference probabilities near 0.5, may provide a proxy.
    \item \textbf{OOD Prompts}: An active learning strategy may focus on prompts or images that are not representative, causing the model to overfit to them. Broad coverage is therefore necessary.
    \item \textbf{Human-in-the-Loop Constraints}: People cannot answer training queries in real time at scale. Active learning may need batched workflows: propose queries, collect answers, update the model, and repeat. Query design, including pairwise comparisons, ratings, or yes/no labels, affects human effort per query.
    \item \textbf{Synthetic or AI Feedback Risks}: If AI systems such as CLIP, LLMs, or vision-language models provide feedback to reduce human load, the diffusion model may align to AI-system biases rather than human intent.
\end{itemize}

\paragraph{Related Work:}

Uehara et al. \cite{uehara2024feedback} directly address feedback efficiency on the algorithmic side by ensuring that each query is used effectively and by exploring on the ``manifold of feasible images'' \cite{uehara2024feedback}. Their bandit formulation suggests a path toward active query selection in which oracle queries are reserved for uncertain cases.

Active learning is well studied in supervised learning, and preference-learning work has considered how to select item pairs that best reveal a user's utility function.

\textbf{RLAIF (reinforcement learning with AI feedback)} in language (like using GPT-4 to critique model outputs instead of humans) has shown promising results, albeit needing careful calibration.

\textbf{Off-policy feedback utilization}: Some works, such as Diffusion-KTO \cite{li2024kto}, exploit existing signals such as user ``likes'' instead of actively querying. Such passive feedback can provide a starting point, with active queries reserved for areas where the signals are absent or ambiguous.

\textit{How can a diffusion model remain safely aligned under adversarial or OOD prompts that attempt to circumvent safety measures?} Many models can be induced to produce disallowed content with carefully phrased prompts, including role-play scenarios or obfuscated language. Unlike classification tasks, where adversarial examples are often pixel-level perturbations, the adversary here is a user choosing a prompt. The challenge is to build a \textbf{robust safety mechanism} that is difficult to bypass.

\paragraph{Key Challenges:}
\begin{itemize}
    \item \textbf{Defining Adversarial Prompts}: The space of prompt attacks is large and open-ended. Evaluation requires representative samples of such prompts.
    \item \textbf{Balancing Content Filtering vs. Generation}: The model should not become overly sensitive and refuse normal requests (false positives), but also not be easily tricked (false negatives of the filter).
    \item \textbf{Integration of Filtering Mechanisms}: External approaches can filter prompts or run outputs through a safety classifier, but integrating these mechanisms into model alignment through RL or fine-tuning is non-trivial.
    \item \textbf{Testing Robustness}: Safety evaluation needs to go beyond static prompt lists, for example through red-team evaluations that actively search for failures.
\end{itemize}

\paragraph{Related Work:}

Focus-N-Fix \cite{xing2025focusfix} improves safety by focusing on problematic output regions, but it does not primarily address adversarial prompts. It fixes image regions associated with known unsafe outputs, which is complementary to prompt-robust alignment.

\textbf{Safety filters}, such as external classifiers or keyword-based prompt filters, often operate outside the model. Simple keyword filters can be brittle because small prompt changes may bypass them.

Adversarial prompt generation can also use language models to generate prompts likely to induce policy-violating images. Related work on adversarial attacks against multimodal models provides useful tools for this setting.

In RL terms, this can be formulated as an adversarial game: one agent searches for prompts that cause safety violations, while the diffusion-model policy updates to avoid them. This resembles adversarial training in classification and generative adversarial learning.

Related language-model work trains on adversarial instruction data, including instructions designed to bypass safety behavior. Techniques from that area, such as refusal behavior under unsafe trigger conditions, can inform image-generation safety alignment.

Robustness should be evaluated through red-team protocols that record both the number and type of successful attacks. An important open question is how to incorporate safety rewards without making normal content generation overly conservative. Potential approaches include weighting schedules that gradually increase adversarial-prompt emphasis and evaluation suites that monitor both unsafe-generation rates and benign-prompt quality.

\textit{Can a diffusion model be continually refined with new feedback and evolving preferences without retraining from scratch and without degrading older preferences?} In deployed settings, user preferences or societal norms may shift, and new content types or edge cases may emerge. This motivates \textbf{continual alignment} methods that learn from new data through periodic updates while preserving previous alignment behavior.

\paragraph{Key Challenges:}
\begin{itemize}
    \item \textbf{Avoiding Forgetting}: As with any continual learning, there is risk the model forgets how to do tasks it was previously good at when learning a new one (or aligning to a new rule).
    \item \textbf{Efficiency of Updates}: Fine-tuning the entire model for every small update is expensive, so local adjustments such as adapter updates may be preferable.
    \item \textbf{Feedback Incorporation}: Deployed systems may receive implicit feedback such as user edits, output rerolls, or explicit ratings. Such signals can be noisy and sparse, requiring careful filtering.
    \item \textbf{Validation and Safety of Continuous Updates}: With continuous change, there is potential to inadvertently break something that was working. Monitoring systems are needed to catch misalignment regressions.
\end{itemize}

\paragraph{Related Work:}

Some recent works have explored continual RLHF, where a model keeps improving with incoming feedback. Many focus on avoiding forgetting by using techniques like Elastic Weight Consolidation or memory replay of old data.

In diffusion models, less work has focused directly on continual learning, but \cite{wang2022fine} suggests using small adaptation networks instead of full model tuning. Modular adapters could support incremental updates, versioning, ensembling, or switching between alignment components.

The concept of \textbf{model editing} from NLP is also relevant: given a specific new fact or rule, one can locally edit model weights to implement it. Analogous approaches for images remain underdeveloped, partly because the meaning of a local visual rule is less direct.

The online RL approach of \cite{uehara2024feedback} is episodic and includes regret guarantees. Similar theoretical tools may help characterize when continual updates converge to a stable solution rather than oscillating or diverging.

\textit{Can reward models or preference models used for alignment be made more interpretable and better aligned with human-understandable concepts?} Current reward models are often black-box neural networks that output a scalar score. If a diffusion model exploits that score, it can be difficult to diagnose why. An interpretable reward model could decompose the score into components such as ``image too dark,'' ``object missing,'' or ``possibly offensive content.'' Such components would help developers debug reward failures and could provide more targeted feedback to the diffusion model, similar in spirit to region-level fixes in Focus-N-Fix.

\paragraph{Key Challenges:}
\begin{itemize}
    \item \textbf{Complexity vs Interpretability}: Human aesthetic preferences and alignment judgments are complex, so interpretable models may oversimplify. Structured criteria can help, but may miss unforeseen aspects.
    \item \textbf{Obtaining Explanations}: Getting human feedback is already hard; getting them to also explain their choices is even harder. But maybe existing datasets have attributes or rationales (some do, albeit small).
    \item \textbf{Using Interpretable Models}: Decision trees or rule-based systems for image preference are unlikely to be accurate enough. A multi-label issue classifier can provide a middle ground, with each label corresponding to a meaningful preference or safety factor.
    \item \textbf{Trust and User Interaction}: Interpretations should be presented to users and model developers in a way that provides evidence about which alignment criteria are being satisfied.
\end{itemize}

\paragraph{Related Work:}
In NLP RLHF, reward-model interpretability is studied as a way to detect biases and spurious correlations. In vision, explainable AI methods can describe image quality or content in human-readable terms, such as identifying blur or missing objects. A related direction is to use textual feedback for images, for example by collecting explanations of why an image is preferred or dispreferred and using those explanations to guide alignment.

Interpretability can be evaluated by measuring whether reward-model factors correlate with human-annotated reasons. The factorized reward must also remain usable for training: if each factor is differentiable, or can be approximated by a differentiable surrogate, methods such as AlignProp can optimize each factor directly; otherwise, multi-objective RL can treat factors as separate objectives. This direction is related to multi-objective alignment, but it focuses on decomposing a broad human-preference objective into auditable reasons.

\subsection*{Summary of Open Directions}
Each of the above research problems addresses a vital aspect of alignment and safety:
\begin{itemize}
    \item Multi-objective alignment extends current methods to handle complex, sometimes conflicting goals in a single model.
    \item Active learning for feedback tackles the data efficiency and scalability of alignment, crucial for real-world deployment where data is limited or costly.
    \item Robust safety alignment proactively defends against misuse, an increasingly important area as generative models become widespread.
    \item Continual alignment adaptation ensures the model remains relevant and aligned over time, akin to maintaining software with updates.
    \item Interpretable reward modeling builds user and stakeholder trust in the alignment process, making the ``black box'' a bit more transparent.
\end{itemize}

These directions are complementary. For instance, an interpretable reward model can be used in active learning (to decide which aspect to query about), and multi-objective alignment naturally will involve interpretability to balance objectives.

\section{\label{sec:conclusion}Conclusion}

Diffusion models hold immense promise for generative AI, but ensuring their outputs remain aligned with human intentions and ethical norms is an ongoing challenge. This survey reviewed recent progress in aligning diffusion models and improving their safety using reinforcement learning, reward modeling, preference optimization, and differentiable reward fine-tuning. We categorized approaches from RLHF to direct preference optimization and analyzed how they interconnect and build upon each other.

While considerable advances have been made, including fine-tuning diffusion models with human feedback to improve prompt adherence \cite{lee2023aligning} and using reward backpropagation to efficiently improve image quality \cite{clark2024draft}, open problems remain before these models can be trusted in sensitive or evolving real-world contexts. The key directions are multi-objective alignment, reduced reliance on large human-labeled datasets, robustness to adversarial prompts, continual learning of preferences, and interpretable reward modeling.

The survey highlights the following priorities for future work:
\begin{itemize}
    \item Unified frameworks for multi-objective diffusion model alignment, validated on simultaneously improving accuracy, aesthetics, and safety without severe trade-offs.
    \item Active learning algorithms that reduce the amount of human feedback needed by a substantial factor, for example by achieving comparable alignment with 50\% fewer queries.
    \item Adversarial training regimens that make diffusion models significantly harder to ``trick'' into producing disallowed content, providing templates for safer deployment.
    \item Methods for maintaining alignment over time, including metrics and techniques to measure and prevent alignment drift as new feedback arrives.
    \item Interpretable reward models that expose the reasons behind preference scores, alongside experiments quantifying the trade-offs between interpretability and performance.
\end{itemize}

Addressing these will push the frontier of not just \emph{what} diffusion models can generate, but \emph{how} and \emph{why} they generate it, aligning their operation with human values and oversight. This is critical as generative models move from lab settings into daily life applications ranging from content creation to design and beyond.

In closing, aligning AI systems with human preferences and ethical principles is a cornerstone of safe AI development. Diffusion models provide a particularly important case study because their outputs are high-dimensional, safety-sensitive, and increasingly deployed in creative and professional workflows. Progress on reward modeling, preference optimization, robust safety alignment, and interpretability for diffusion models is therefore likely to inform alignment strategies for other modalities and future generations of generative models.

\bibliographystyle{unsrtnat}

\bibliography{references}

@misc{lee2023aligning,
      title={Aligning Text-to-Image Models using Human Feedback}, 
      author={Kimin Lee and Hao Liu and Moonkyung Ryu and Olivia Watkins and Yuqing Du and Craig Boutilier and Pieter Abbeel and Mohammad Ghavamzadeh and Shixiang Shane Gu},
      year={2023},
      note={arXiv: 2302.12192},
      archivePrefix={arXiv},
      primaryClass={cs.LG},
      url={https://arxiv.org/abs/2302.12192}, 
}

@inproceedings{fan2023dpok,
title={Reinforcement Learning for Fine-tuning Text-to-Image Diffusion Models},
author={Ying Fan and Olivia Watkins and Yuqing Du and Hao Liu and Moonkyung Ryu and Craig Boutilier and Pieter Abbeel and Mohammad Ghavamzadeh and Kangwook Lee and Kimin Lee},
booktitle={Thirty-seventh Conference on Neural Information Processing Systems},
year={2023},
url={https://openreview.net/forum?id=8OTPepXzeh}
}

@misc{prabhudesai2023alignprop,
      title={Aligning Text-to-Image Diffusion Models with Reward Backpropagation}, 
      author={Mihir Prabhudesai and Anirudh Goyal and Deepak Pathak and Katerina Fragkiadaki},
      year={2024},
      eprint={2310.03739},
      archivePrefix={arXiv},
      primaryClass={cs.CV},
      url={https://arxiv.org/abs/2310.03739}, 
}

@INPROCEEDINGS{deng2024prdp,
  author={Deng, Fei and Wang, Qifei and Wei, Wei and Hou, Tingbo and Grundmann, Matthias},
  booktitle={2024 IEEE/CVF Conference on Computer Vision and Pattern Recognition (CVPR)}, 
  title={PRDP: Proximal Reward Difference Prediction for Large-Scale Reward Finetuning of Diffusion Models}, 
  year={2024},
  volume={},
  number={},
  pages={7423-7433},
  doi={10.1109/CVPR52733.2024.00709}
}

@article{zhang2024aligningfewstepdiffusionmodels,
   title={Aligning Few-Step Diffusion Models with Dense Reward Difference Learning},
   ISSN={1939-3539},
   url={http://dx.doi.org/10.1109/TPAMI.2026.3665753},
   DOI={10.1109/tpami.2026.3665753},
   journal={IEEE Transactions on Pattern Analysis and Machine Intelligence},
   publisher={Institute of Electrical and Electronics Engineers (IEEE)},
   author={Zhang, Ziyi and Shen, Li and Zhang, Sen and Ye, Deheng and Luo, Yong and Shi, Miaojing and Shan, Dongjing and Du, Bo and Tao, Dacheng},
   year={2026},
   pages={1–12} }

@inproceedings{yang2024dense,
author = {Yang, Shentao and Chen, Tianqi and Zhou, Mingyuan},
title = {A dense reward view on aligning text-to-image diffusion with preference},
year = {2024},
publisher = {JMLR.org},
booktitle = {Proceedings of the 41st International Conference on Machine Learning},
articleno = {2310},
numpages = {35},
location = {Vienna, Austria},
series = {ICML'24}
}

@inproceedings{xu2023imagereward,
author = {Xu, Jiazheng and Liu, Xiao and Wu, Yuchen and Tong, Yuxuan and Li, Qinkai and Ding, Ming and Tang, Jie and Dong, Yuxiao},
title = {ImageReward: learning and evaluating human preferences for text-to-image generation},
year = {2023},
publisher = {Curran Associates Inc.},
address = {Red Hook, NY, USA},
booktitle = {Proceedings of the 37th International Conference on Neural Information Processing Systems},
articleno = {700},
numpages = {33},
location = {New Orleans, LA, USA},
series = {NIPS '23}
}

@misc{kirillov2023segment,
      title={Segment Anything}, 
      author={Alexander Kirillov and Eric Mintun and Nikhila Ravi and Hanzi Mao and Chloe Rolland and Laura Gustafson and Tete Xiao and Spencer Whitehead and Alexander C. Berg and Wan-Yen Lo and Piotr Dollár and Ross Girshick},
      year={2023},
      note={arXiv: 2304.02643},
      archivePrefix={arXiv},
      primaryClass={cs.CV},
      url={https://arxiv.org/abs/2304.02643}, 
}

@inproceedings{ouyang2022training,
author = {Ouyang, Long and Wu, Jeff and Jiang, Xu and Almeida, Diogo and Wainwright, Carroll L. and Mishkin, Pamela and Zhang, Chong and Agarwal, Sandhini and Slama, Katarina and Ray, Alex and Schulman, John and Hilton, Jacob and Kelton, Fraser and Miller, Luke and Simens, Maddie and Askell, Amanda and Welinder, Peter and Christiano, Paul and Leike, Jan and Lowe, Ryan},
title = {Training language models to follow instructions with human feedback},
year = {2022},
isbn = {9781713871088},
publisher = {Curran Associates Inc.},
address = {Red Hook, NY, USA},
booktitle = {Proceedings of the 36th International Conference on Neural Information Processing Systems},
articleno = {2011},
numpages = {15},
location = {New Orleans, LA, USA},
series = {NIPS '22}
}

@inproceedings{christiano2017deep,
author = {Christiano, Paul F. and Leike, Jan and Brown, Tom B. and Martic, Miljan and Legg, Shane and Amodei, Dario},
title = {Deep reinforcement learning from human preferences},
year = {2017},
isbn = {9781510860964},
publisher = {Curran Associates Inc.},
address = {Red Hook, NY, USA},
booktitle = {Proceedings of the 31st International Conference on Neural Information Processing Systems},
pages = {4302–4310},
numpages = {9},
location = {Long Beach, California, USA},
series = {NIPS'17}
}

@inproceedings{rafailov2023dpo,
author = {Rafailov, Rafael and Sharma, Archit and Mitchell, Eric and Ermon, Stefano and Manning, Christopher D. and Finn, Chelsea},
title = {Direct preference optimization: your language model is secretly a reward model},
year = {2023},
publisher = {Curran Associates Inc.},
address = {Red Hook, NY, USA},
booktitle = {Proceedings of the 37th International Conference on Neural Information Processing Systems},
articleno = {2338},
numpages = {14},
location = {New Orleans, LA, USA},
series = {NIPS '23}
}

@misc{wu2023humanpreferencescorev2,
      title={Human Preference Score v2: A Solid Benchmark for Evaluating Human Preferences of Text-to-Image Synthesis}, 
      author={Xiaoshi Wu and Yiming Hao and Keqiang Sun and Yixiong Chen and Feng Zhu and Rui Zhao and Hongsheng Li},
      year={2023},
      eprint={2306.09341},
      archivePrefix={arXiv},
      primaryClass={cs.CV},
      url={https://arxiv.org/abs/2306.09341}, 
}

@INPROCEEDINGS{xing2025focusfix,
  author={Xing, Xiaoying and Saha, Avinab and He, Junfeng and Hao, Susan and Vicol, Paul and Ryu, Moonkyung and Li, Gang and Singla, Sahil and Young, Sarah and Li, Yinxiao and Yang, Feng and Ramachandran, Deepak},
  booktitle={2025 IEEE/CVF Conference on Computer Vision and Pattern Recognition (CVPR)}, 
  title={Focus-N-Fix: Region-Aware Fine-Tuning for Text-to-Image Generation}, 
  year={2025},
  volume={},
  number={},
  pages={18486-18496},
  doi={10.1109/CVPR52734.2025.01723}}

@inproceedings{zhang2024rewardoveropt,
author = {Zhang, Ziyi and Zhang, Sen and Zhan, Yibing and Luo, Yong and Wen, Yonggang and Tao, Dacheng},
title = {Confronting reward overoptimization for diffusion models: a perspective of inductive and primacy biases},
year = {2024},
publisher = {JMLR.org},
booktitle = {Proceedings of the 41st International Conference on Machine Learning},
articleno = {2500},
numpages = {18},
location = {Vienna, Austria},
series = {ICML'24}
}

@misc{gupta2025simpleeffectivereinforcementlearning,
      title={A Simple and Effective Reinforcement Learning Method for Text-to-Image Diffusion Fine-tuning}, 
      author={Shashank Gupta and Chaitanya Ahuja and Tsung-Yu Lin and Sreya Dutta Roy and Harrie Oosterhuis and Maarten de Rijke and Satya Narayan Shukla},
      year={2026},
      note={arXiv: 2503.00897},
      archivePrefix={arXiv},
      primaryClass={cs.LG},
      url={https://arxiv.org/abs/2503.00897}, 
}

@inproceedings{schuhmann2022laion,
author = {Schuhmann, Christoph and Beaumont, Romain and Vencu, Richard and Gordon, Cade and Wightman, Ross and Cherti, Mehdi and Coombes, Theo and Katta, Aarush and Mullis, Clayton and Wortsman, Mitchell and Schramowski, Patrick and Kundurthy, Srivatsa and Crowson, Katherine and Schmidt, Ludwig and Kaczmarczyk, Robert and Jitsev, Jenia},
title = {LAION-5B: an open large-scale dataset for training next generation image-text models},
year = {2022},
isbn = {9781713871088},
publisher = {Curran Associates Inc.},
address = {Red Hook, NY, USA},
booktitle = {Proceedings of the 36th International Conference on Neural Information Processing Systems},
articleno = {1833},
numpages = {17},
location = {New Orleans, LA, USA},
series = {NIPS '22}
}

@inproceedings{li2022blip,
  title={Blip: Bootstrapping language-image pre-training for unified vision-language understanding and generation},
  author={Li, Junnan and Li, Dongxu and Xiong, Caiming and Hoi, Steven},
  booktitle={International conference on machine learning},
  pages={12888--12900},
  year={2022},
  organization={PMLR}
}

@inproceedings{heusel2017gans,
author = {Heusel, Martin and Ramsauer, Hubert and Unterthiner, Thomas and Nessler, Bernhard and Hochreiter, Sepp},
title = {GANs trained by a two time-scale update rule converge to a local nash equilibrium},
year = {2017},
isbn = {9781510860964},
publisher = {Curran Associates Inc.},
address = {Red Hook, NY, USA},
booktitle = {Proceedings of the 31st International Conference on Neural Information Processing Systems},
pages = {6629–6640},
numpages = {12},
location = {Long Beach, California, USA},
series = {NIPS'17}
}

@misc{radford2021learning,
      title={Learning Transferable Visual Models From Natural Language Supervision}, 
      author={Alec Radford and Jong Wook Kim and Chris Hallacy and Aditya Ramesh and Gabriel Goh and Sandhini Agarwal and Girish Sastry and Amanda Askell and Pamela Mishkin and Jack Clark and Gretchen Krueger and Ilya Sutskever},
      year={2021},
      eprint={2103.00020},
      archivePrefix={arXiv},
      primaryClass={cs.CV},
      url={https://arxiv.org/abs/2103.00020}, 
}

@inproceedings{uehara2024feedback,
author = {Uehara, Masatoshi and Zhao, Yulai and Black, Kevin and Hajiramezanali, Ehsan and Scalia, Gabriele and Diamant, Nathaniel Lee and Tseng, Alex M and Levine, Sergey and Biancalani, Tommaso},
title = {Feedback efficient online fine-tuning of diffusion models},
year = {2024},
publisher = {JMLR.org},
booktitle = {Proceedings of the 41st International Conference on Machine Learning},
articleno = {1999},
numpages = {27},
location = {Vienna, Austria},
series = {ICML'24}
}

@inproceedings{ho2020ddpm,
author = {Ho, Jonathan and Jain, Ajay and Abbeel, Pieter},
title = {Denoising diffusion probabilistic models},
year = {2020},
isbn = {9781713829546},
publisher = {Curran Associates Inc.},
address = {Red Hook, NY, USA},
booktitle = {Proceedings of the 34th International Conference on Neural Information Processing Systems},
articleno = {574},
numpages = {12},
location = {Vancouver, BC, Canada},
series = {NIPS '20}
}

@misc{ramesh2022hierarchical,
      title={Hierarchical Text-Conditional Image Generation with CLIP Latents}, 
      author={Aditya Ramesh and Prafulla Dhariwal and Alex Nichol and Casey Chu and Mark Chen},
      year={2022},
      note={arXiv: 2204.06125},
      archivePrefix={arXiv},
      primaryClass={cs.CV},
      url={https://arxiv.org/abs/2204.06125}, 
}

@misc{ramesh2022dalle2,
      title={Hierarchical Text-Conditional Image Generation with CLIP Latents}, 
      author={Aditya Ramesh and Prafulla Dhariwal and Alex Nichol and Casey Chu and Mark Chen},
      year={2022},
      eprint={2204.06125},
      archivePrefix={arXiv},
      primaryClass={cs.CV},
      url={https://arxiv.org/abs/2204.06125}, 
}

@inproceedings{wang2022fine,
title={Diffusion Policies as an Expressive Policy Class for Offline Reinforcement Learning},
author={Zhendong Wang and Jonathan J Hunt and Mingyuan Zhou},
booktitle={The Eleventh International Conference on Learning Representations },
year={2023},
url={https://openreview.net/forum?id=AHvFDPi-FA}
}

@inproceedings{saharia2022imagen,
author = {Saharia, Chitwan and Chan, William and Saxena, Saurabh and Lit, Lala and Whang, Jay and Denton, Emily and Ghasemipour, Seyed Kamyar Seyed and Ayan, Burcu Karagol and Mahdavi, S. Sara and Gontijo-Lopes, Raphael and Salimans, Tim and Ho, Jonathan and Fleet, David J and Norouzi, Mohammad},
title = {Photorealistic text-to-image diffusion models with deep language understanding},
year = {2022},
isbn = {9781713871088},
publisher = {Curran Associates Inc.},
address = {Red Hook, NY, USA},
booktitle = {Proceedings of the 36th International Conference on Neural Information Processing Systems},
articleno = {2643},
numpages = {16},
location = {New Orleans, LA, USA},
series = {NIPS '22}
}

@misc{nichol2022glide,
      title={GLIDE: Towards Photorealistic Image Generation and Editing with Text-Guided Diffusion Models}, 
      author={Alex Nichol and Prafulla Dhariwal and Aditya Ramesh and Pranav Shyam and Pamela Mishkin and Bob McGrew and Ilya Sutskever and Mark Chen},
      year={2022},
      note={arXiv: 2112.10741},
      archivePrefix={arXiv},
      primaryClass={cs.CV},
      url={https://arxiv.org/abs/2112.10741}, 
}

@misc{nichol2021improved,
      title={Improved Denoising Diffusion Probabilistic Models}, 
      author={Alex Nichol and Prafulla Dhariwal},
      year={2021},
      note={arXiv: 2102.09672},
      archivePrefix={arXiv},
      primaryClass={cs.LG},
      url={https://arxiv.org/abs/2102.09672}, 
}

@misc{kirstain2023pickapic,
      title={Pick-a-Pic: An Open Dataset of User Preferences for Text-to-Image Generation}, 
      author={Yuval Kirstain and Adam Polyak and Uriel Singer and Shahbuland Matiana and Joe Penna and Omer Levy},
      year={2023},
      note={arXiv: 2305.01569},
      archivePrefix={arXiv},
      primaryClass={cs.CV},
      url={https://arxiv.org/abs/2305.01569}, 
}

@InProceedings{yang2024d3po,
    author    = {Yang, Kai and Tao, Jian and Lyu, Jiafei and Ge, Chunjiang and Chen, Jiaxin and Shen, Weihan and Zhu, Xiaolong and Li, Xiu},
    title     = {Using Human Feedback to Fine-tune Diffusion Models without Any Reward Model},
    booktitle = {Proceedings of the IEEE/CVF Conference on Computer Vision and Pattern Recognition (CVPR)},
    month     = {June},
    year      = {2024},
    pages     = {8941-8951}
}

@inproceedings{li2024diffusionkto,
title={Aligning Diffusion Models by Optimizing Human Utility},
author={Shufan Li and Konstantinos Kallidromitis and Akash Gokul and Yusuke Kato and Kazuki Kozuka},
booktitle={The Thirty-eighth Annual Conference on Neural Information Processing Systems},
year={2024},
url={https://openreview.net/forum?id=MTMShU5QaC}
}

@misc{li2024kto,
      title={Aligning Diffusion Models by Optimizing Human Utility}, 
      author={Shufan Li and Konstantinos Kallidromitis and Akash Gokul and Yusuke Kato and Kazuki Kozuka},
      year={2024},
      eprint={2404.04465},
      archivePrefix={arXiv},
      primaryClass={cs.CV},
      url={https://arxiv.org/abs/2404.04465}, 
}

@misc{xing2025focus,
      title={Focus-N-Fix: Region-Aware Fine-Tuning for Text-to-Image Generation}, 
      author={Xiaoying Xing and Avinab Saha and Junfeng He and Susan Hao and Paul Vicol and Moonkyung Ryu and Gang Li and Sahil Singla and Sarah Young and Yinxiao Li and Feng Yang and Deepak Ramachandran},
      year={2024},
      note={arXiv: 2501.06481},
      archivePrefix={arXiv},
      primaryClass={cs.CV},
      url={https://arxiv.org/abs/2501.06481}, 
}

@misc{liu2024safetydposcalablesafetyalignment,
      title={AlignGuard: Scalable Safety Alignment for Text-to-Image Generation}, 
      author={Runtao Liu and I Chieh Chen and Jindong Gu and Jipeng Zhang and Renjie Pi and Qifeng Chen and Philip Torr and Ashkan Khakzar and Fabio Pizzati},
      year={2025},
      note={arXiv: 2412.10493},
      archivePrefix={arXiv},
      primaryClass={cs.CV},
      url={https://arxiv.org/abs/2412.10493}, 
}

@misc{liao2025helmhumanpreferredexplorationlanguage,
      title={HELM: Human-Preferred Exploration with Language Models}, 
      author={Shuhao Liao and Xuxin Lv and Yuhong Cao and Jeric Lew and Wenjun Wu and Guillaume Sartoretti},
      year={2025},
      note={arXiv: 2503.07006},
      archivePrefix={arXiv},
      primaryClass={cs.RO},
      url={https://arxiv.org/abs/2503.07006}, 
}

@inproceedings{zhu2025dspo,
title={{DSPO}: Direct Score Preference Optimization for Diffusion Model Alignment},
author={Huaisheng Zhu and Teng Xiao and Vasant G Honavar},
booktitle={The Thirteenth International Conference on Learning Representations},
year={2025},
url={https://openreview.net/forum?id=xyfb9HHvMe}
}

@misc{liu2024alignmentdiffusionmodelsfundamentals,
      title={Alignment of Diffusion Models: Fundamentals, Challenges, and Future}, 
      author={Buhua Liu and Shitong Shao and Bao Li and Lichen Bai and Zhiqiang Xu and Haoyi Xiong and James Kwok and Sumi Helal and Zeke Xie},
      year={2026},
      note={arXiv: 2409.07253},
      archivePrefix={arXiv},
      primaryClass={cs.LG},
      url={https://arxiv.org/abs/2409.07253}, 
}

@misc{wallace2023diffusionmodelalignmentusing,
      title={Diffusion Model Alignment Using Direct Preference Optimization}, 
      author={Bram Wallace and Meihua Dang and Rafael Rafailov and Linqi Zhou and Aaron Lou and Senthil Purushwalkam and Stefano Ermon and Caiming Xiong and Shafiq Joty and Nikhil Naik},
      year={2023},
      note={arXiv: 2311.12908},
      archivePrefix={arXiv},
      primaryClass={cs.CV},
      url={https://arxiv.org/abs/2311.12908}, 
}

@article{Kingma_2019,
   title={An Introduction to Variational Autoencoders},
   volume={12},
   ISSN={1935-8245},
   url={http://dx.doi.org/10.1561/2200000056},
   DOI={10.1561/2200000056},
   number={4},
   journal={Foundations and Trends® in Machine Learning},
   publisher={Emerald},
   author={Kingma, Diederik P. and Welling, Max},
   year={2019},
   month=Nov, pages={307–392} }

@misc{clark2024draft,
      title={Directly Fine-Tuning Diffusion Models on Differentiable Rewards}, 
      author={Kevin Clark and Paul Vicol and Kevin Swersky and David J Fleet},
      year={2024},
      note={arXiv: 2309.17400},
      archivePrefix={arXiv},
      primaryClass={cs.CV},
      url={https://arxiv.org/abs/2309.17400}, 
}

@inproceedings{lee2024rlaifvsrlhfscaling,
author = {Lee, Harrison and Phatale, Samrat and Mansoor, Hassan and Mesnard, Thomas and Ferret, Johan and Lu, Kellie and Bishop, Colton and Hall, Ethan and Carbune, Victor and Rastogi, Abhinav and Prakash, Sushant},
title = {RLAIF vs. RLHF: scaling reinforcement learning from human feedback with AI feedback},
year = {2024},
publisher = {JMLR.org},
booktitle = {Proceedings of the 41st International Conference on Machine Learning},
articleno = {1071},
numpages = {28},
location = {Vienna, Austria},
series = {ICML'24}
}

@misc{bordes2024introductionvisionlanguagemodeling,
      title={An Introduction to Vision-Language Modeling}, 
      author={Florian Bordes and Richard Yuanzhe Pang and Anurag Ajay and Alexander C. Li and Adrien Bardes and Suzanne Petryk and Oscar Mañas and Zhiqiu Lin and Anas Mahmoud and Bargav Jayaraman and Mark Ibrahim and Melissa Hall and Yunyang Xiong and Jonathan Lebensold and Candace Ross and Srihari Jayakumar and Chuan Guo and Diane Bouchacourt and Haider Al-Tahan and Karthik Padthe and Vasu Sharma and Hu Xu and Xiaoqing Ellen Tan and Megan Richards and Samuel Lavoie and Pietro Astolfi and Reyhane Askari Hemmat and Jun Chen and Kushal Tirumala and Rim Assouel and Mazda Moayeri and Arjang Talattof and Kamalika Chaudhuri and Zechun Liu and Xilun Chen and Quentin Garrido and Karen Ullrich and Aishwarya Agrawal and Kate Saenko and Asli Celikyilmaz and Vikas Chandra},
      year={2024},
      note={arXiv: 2405.17247},
      archivePrefix={arXiv},
      primaryClass={cs.LG},
      url={https://arxiv.org/abs/2405.17247}, 
}

@inproceedings{dhariwal2021diffusion,
author = {Dhariwal, Prafulla and Nichol, Alex},
title = {Diffusion models beat GANs on image synthesis},
year = {2021},
isbn = {9781713845393},
publisher = {Curran Associates Inc.},
address = {Red Hook, NY, USA},
booktitle = {Proceedings of the 35th International Conference on Neural Information Processing Systems},
articleno = {672},
numpages = {15},
series = {NIPS '21}
}

@misc{jia2025lasro,
      title={Reward Fine-Tuning Two-Step Diffusion Models via Learning Differentiable Latent-Space Surrogate Reward}, 
      author={Zhiwei Jia and Yuesong Nan and Huixi Zhao and Gengdai Liu},
      year={2025},
      eprint={2411.15247},
      archivePrefix={arXiv},
      primaryClass={cs.LG},
      url={https://arxiv.org/abs/2411.15247}, 
}

@INPROCEEDINGS {rombach2022high,
author = { Rombach, Robin and Blattmann, Andreas and Lorenz, Dominik and Esser, Patrick and Ommer, Bjorn },
booktitle = { 2022 IEEE/CVF Conference on Computer Vision and Pattern Recognition (CVPR) },
title = {{ High-Resolution Image Synthesis with Latent Diffusion Models }},
year = {2022},
volume = {},
ISSN = {},
pages = {10674-10685},
doi = {10.1109/CVPR52688.2022.01042},
url = {https://doi.ieeecomputersociety.org/10.1109/CVPR52688.2022.01042},
publisher = {IEEE Computer Society},
address = {Los Alamitos, CA, USA},
month =Jun}

@misc{smith2024diffusiondpo,
      title={Diffusion Model Alignment Using Direct Preference Optimization}, 
      author={Bram Wallace and Meihua Dang and Rafael Rafailov and Linqi Zhou and Aaron Lou and Senthil Purushwalkam and Stefano Ermon and Caiming Xiong and Shafiq Joty and Nikhil Naik},
      year={2023},
      eprint={2311.12908},
      archivePrefix={arXiv},
      primaryClass={cs.CV},
      url={https://arxiv.org/abs/2311.12908}, 
}

@misc{lee2024diffusionkto,
      title={Aligning Diffusion Models by Optimizing Human Utility}, 
      author={Shufan Li and Konstantinos Kallidromitis and Akash Gokul and Yusuke Kato and Kazuki Kozuka},
      year={2024},
      eprint={2404.04465},
      archivePrefix={arXiv},
      primaryClass={cs.CV},
      url={https://arxiv.org/abs/2404.04465}, 
}

@INPROCEEDINGS{zhang2024surrogate,
  author={Jia, Zhiwei and Nan, Yuesong and Zhao, Huixi and Liu, Gengdai},
  booktitle={2025 IEEE/CVF Conference on Computer Vision and Pattern Recognition (CVPR)}, 
  title={Reward Fine-Tuning Two-Step Diffusion Models via Learning Differentiable Latent-Space Surrogate Reward}, 
  year={2025},
  volume={},
  number={},
  pages={12912-12922},
  doi={10.1109/CVPR52734.2025.01205}}

@misc{wang2024prdp,
      title={PRDP: Proximal Reward Difference Prediction for Large-Scale Reward Finetuning of Diffusion Models}, 
      author={Fei Deng and Qifei Wang and Wei Wei and Matthias Grundmann and Tingbo Hou},
      year={2024},
      note={arXiv: 2402.08714},
      archivePrefix={arXiv},
      primaryClass={cs.LG},
      url={https://arxiv.org/abs/2402.08714}, 
}

@misc{song2021scorebasedgenerativemodelingstochastic,
      title={Score-Based Generative Modeling through Stochastic Differential Equations}, 
      author={Yang Song and Jascha Sohl-Dickstein and Diederik P. Kingma and Abhishek Kumar and Stefano Ermon and Ben Poole},
      year={2021},
      eprint={2011.13456},
      archivePrefix={arXiv},
      primaryClass={cs.LG},
      url={https://arxiv.org/abs/2011.13456}, 
}

@inproceedings{austin2021structured,
author = {Austin, Jacob and Johnson, Daniel D. and Ho, Jonathan and Tarlow, Daniel and van den Berg, Rianne},
title = {Structured denoising diffusion models in discrete state-spaces},
year = {2021},
isbn = {9781713845393},
publisher = {Curran Associates Inc.},
address = {Red Hook, NY, USA},
booktitle = {Proceedings of the 35th International Conference on Neural Information Processing Systems},
articleno = {1376},
numpages = {13},
series = {NIPS '21}
}

@inproceedings{li2022diffusionlm,
title={Diffusion-{LM} Improves Controllable Text Generation},
author={Xiang Lisa Li and John Thickstun and Ishaan Gulrajani and Percy Liang and Tatsunori Hashimoto},
booktitle={Advances in Neural Information Processing Systems},
editor={Alice H. Oh and Alekh Agarwal and Danielle Belgrave and Kyunghyun Cho},
year={2022},
url={https://openreview.net/forum?id=3s9IrEsjLyk}
}

@inproceedings{gong2023diffuseq,
title={DiffuSeq: Sequence to Sequence Text Generation with Diffusion Models},
author={Shansan Gong and Mukai Li and Jiangtao Feng and Zhiyong Wu and Lingpeng Kong},
booktitle={The Eleventh International Conference on Learning Representations },
year={2023},
url={https://openreview.net/forum?id=jQj-_rLVXsj}
}

@inproceedings{lou2024sedd,
author = {Lou, Aaron and Meng, Chenlin and Ermon, Stefano},
title = {Discrete diffusion modeling by estimating the ratios of the data distribution},
year = {2024},
publisher = {JMLR.org},
booktitle = {Proceedings of the 41st International Conference on Machine Learning},
articleno = {1333},
numpages = {30},
location = {Vienna, Austria},
series = {ICML'24}
}

@inproceedings{sahoo2024mdlm,
author = {Sahoo, Subham Sekhar and Arriola, Marianne and Schiff, Yair and Gokaslan, Aaron and Marroquin, Edgar and Chiu, Justin T and Rush, Alexander and Kuleshov, Volodymyr},
title = {Simple and effective masked diffusion language models},
year = {2024},
isbn = {9798331314385},
publisher = {Curran Associates Inc.},
address = {Red Hook, NY, USA},
booktitle = {Proceedings of the 38th International Conference on Neural Information Processing Systems},
articleno = {4135},
numpages = {49},
location = {Vancouver, BC, Canada},
series = {NIPS '24}
}

@misc{nie2025llada,
  title={Large Language Diffusion Models},
  author={Nie, Shen and Zhu, Fengqi and You, Zebin and Zhang, Xiaolu and Ou, Jingyang and Hu, Jun and Zhou, Jun and Lin, Yankai and Wen, Ji-Rong and Li, Chongxuan},
  year={2025},
  eprint={2502.09992},
  archivePrefix={arXiv},
  primaryClass={cs.CL},
  url={https://arxiv.org/abs/2502.09992}
}

@misc{zhu2025llada15,
  title={{LLaDA} 1.5: Variance-Reduced Preference Optimization for Large Language Diffusion Models},
  author={Zhu, Fengqi and Wang, Rongzhen and Nie, Shen and Zhang, Xiaolu and Wu, Chunwei and Hu, Jun and Zhou, Jun and Chen, Jianfei and Lin, Yankai and Wen, Ji-Rong and Li, Chongxuan},
  year={2025},
  eprint={2505.19223},
  archivePrefix={arXiv},
  primaryClass={cs.LG},
  url={https://arxiv.org/abs/2505.19223}
}

@misc{xue2025dancegrpo,
  title={{DanceGRPO}: Unleashing {GRPO} on Visual Generation},
  author={Xue, Zeyue and Wu, Jie and Gao, Yu and Kong, Fangyuan and Zhu, Lingting and Chen, Mengzhao and Liu, Zhiheng and Liu, Wei and Guo, Qiushan and Huang, Weilin and Luo, Ping},
  year={2025},
  eprint={2505.07818},
  archivePrefix={arXiv},
  primaryClass={cs.CV},
  url={https://arxiv.org/abs/2505.07818}
}

@misc{liu2025flowgrpo,
  title={{Flow-GRPO}: Training Flow Matching Models via Online {RL}},
  author={Liu, Jie and Liu, Gongye and Liang, Jiajun and Li, Yangguang and Liu, Jiaheng and Wang, Xintao and Wan, Pengfei and Zhang, Di and Ouyang, Wanli},
  year={2025},
  eprint={2505.05470},
  archivePrefix={arXiv},
  primaryClass={cs.CV},
  url={https://arxiv.org/abs/2505.05470}
}

@misc{li2025branchgrpo,
  title={{BranchGRPO}: Stable and Efficient {GRPO} with Structured Branching in Diffusion Models},
  author={Li, Yuming and Wang, Yikai and Zhu, Yuying and Zhao, Zhongyu and Lu, Ming and She, Qi and Zhang, Shanghang},
  year={2025},
  eprint={2509.06040},
  archivePrefix={arXiv},
  primaryClass={cs.CV},
  url={https://arxiv.org/abs/2509.06040}
}

@misc{li2026aegpo,
  title={{AEGPO}: Adaptive Entropy-Guided Policy Optimization for Diffusion Models},
  author={Li, Yuming and Li, Qingyu and Bai, Chengyu and Luo, Xiangyang and Xue, Zeyue and Qin, Wenyu and Wang, Meng and Wang, Yikai and Zhang, Shanghang},
  year={2026},
  eprint={2602.06825},
  archivePrefix={arXiv},
  primaryClass={cs.LG},
  url={https://arxiv.org/abs/2602.06825}
}

@misc{dang2025personalizedpreference,
      title={Personalized Preference Fine-tuning of Diffusion Models}, 
      author={Meihua Dang and Anikait Singh and Linqi Zhou and Stefano Ermon and Jiaming Song},
      year={2025},
      note={arXiv: 2501.06655},
      archivePrefix={arXiv},
      primaryClass={cs.LG},
      url={https://arxiv.org/abs/2501.06655}, 
}

@inproceedings{dunlop2025personalizedediting,
title={Personalized Image Editing in Text-to-Image Diffusion Models via Collaborative Direct Preference Optimization},
author={Connor Dunlop and Matthew Zheng and Kavana Venkatesh and Pinar Yanardag},
booktitle={The Thirty-ninth Annual Conference on Neural Information Processing Systems},
year={2026},
url={https://openreview.net/forum?id=BBZEcVu1nA}
}

\end{document}